%% file: Yildirim-2022-IS.tex
\documentclass[a4paper,jou,natbib]{apa7}

\usepackage[english]{babel}
\usepackage[utf8]{inputenc}
\usepackage{amsmath}
\usepackage{subcaption}
\usepackage{graphicx}
\usepackage{dblfloatfix}
\usepackage{color}
\graphicspath{{figures/}} % Graphics will be here

\title{Learning Social Navigation from Demonstrations with Conditional Neural Processes}
\shorttitle{Social Navigation with Conditional Neural Processes} % max 55 char running title
\leftheader{Yildirim and Ugur}
\authorsnames{Yigit Yildirim, Emre Ugur}
\authorsaffiliations{{Computer Engineering Department, Bogazici University, Istanbul, Turkey}}

\abstract{Sociability is essential for modern robots to increase their acceptability in human environments. Traditional techniques use manually engineered utility functions inspired by observing pedestrian behaviors to achieve social navigation. However, social aspects of navigation are diverse, changing across different types of environments, societies, and population densities, making it unrealistic to use hand-crafted techniques in each domain. This paper presents a data-driven navigation architecture that uses state-of-the-art neural architectures, namely Conditional Neural Processes, to learn global and local controllers of the mobile robot from observations. Additionally, we leverage a state-of-the-art, deep prediction mechanism to detect situations not similar to the trained ones, where reactive controllers step in to ensure safe navigation. Our results demonstrate that the proposed framework can successfully carry out navigation tasks regarding social norms in the data. Further, we showed that our system produces fewer personal-zone violations, causing less discomfort.}


\begin{document}
\maketitle

\section{Introduction}

Researchers have been studying mobile robot navigation for decades. Many notable techniques have been proposed in this area over the years, such as \citep{burgard1999ExperiencesWA,thrun2000probabilistic, mobot}, where safety and robustness features have been prioritized. In other words, the principal driving factor behind the development in this field has been collision avoidance \citep{dwa}. On the other hand, as humans start to share their environment with robots, new requirements for mobile robot navigation have emerged.

Physical and mental aspects of safety were separately evaluated in \cite{nonaka2004evaluation}. This separation reveals the need to question the psychological efficiency of the navigation systems of mobile robots. To ensure the smooth integration of robots into human environments, these systems must be social and as natural and understandable to humans as they are safe.

% \remyigit{Social navigation corresponds to navigation that adheres to the social rules of the people \citep{kretzschmar2016socially}. People move socially in their environment. Therefore, it is only natural to imitate their behavior in robots. However, replicating human behavior introduces new constraints that the robot navigation systems must meet. Firstly, human navigation behavior must be formally defined. Then, this formalism must be applied to robot navigation systems.}

\cite{kruse2013human} defines social navigation as navigating the robot in such a way that minimizes the annoyance that its motion produces. Efforts to decrease the annoyance or anxiety of the pedestrians interacting with a navigating robot can be included in the social navigation domain. To this end, the majority of the studies in the literature target to increase the comfortableness of the interaction. \cite{fong2003survey} asserts that people find it more comforting to interact with machines same the same way they interact with other people. Therefore, many researchers have aimed to decrease the discomfort that robot navigation generates by replicating human navigation with mobile robots.

The studies in the literature that aim to imitate human behavior in mobile robot navigation fall into two categories: manually coded controllers and learning-based ones. Manually coded controllers rely on hand-crafted optimization functions to resemble the motion of robots to that of humans. One of the notable studies of this category is the Social Force Model (SFM) \citep{helbing1995social}. Based on behavioral techniques from the social sciences, SFM suggests that pedestrians move under the effect of specific abstract forces, just like particles in an electric field. While the navigational goal attracts the pedestrian, obstacles and other people exert repulsive forces. Despite its wide application \citep{zanlungo2011social, ferrer2013robot, farina2017walking}, some researchers argue that hand-crafted models have limited applicability in controlled environments \citep{vasquez2014inverse} and that they are not general and applicable to different, varying social environments, especially during avoidance maneuvers \citep{kretzschmar2016socially}. In real-world scenarios, social compliance of robot navigation requires adaptability. \cite{kuderer2012feature} proposed the use of data-driven approaches to create such adaptive controllers. Researchers have used numerous machine learning algorithms to create better adaptive, socially compliant navigation frameworks. One of the most popular algorithms is Inverse Reinforcement Learning (IRL) \citep{kuderer2012feature, kitani2012activity,vasquez2014inverse,kim2016socially}. Given perfect expert demonstrations, IRL attempts to identify the underlying reward structure, which can be used by any Reinforcement Learning (RL) algorithm to create a human-aware navigation policy. The advantage of this approach is that the reward function is not manually determined but is a linear combination of a set of predefined features. However, the linearity assumption is considered a strong assumption in \cite{wulfmeier2015maximum}.

Nonlinear rewards can better describe complex behaviors in many real-world problems \citep{levine2011nonlinear}. Researchers have been using deep learning techniques to leverage this potential in social navigation. In \cite{chen2017}, Deep Reinforcement Learning was used to obtain a socially plausible navigation policy. As with other RL approaches, this procedure relies on a predefined reward. Similarly, \cite{wulfmeier2015maximum} extracted nonlinear rewards, assuming that the features shaping the reward function are known. On the other hand, Imitation Learning attempts to learn policies directly from the data, relaxing assumptions about the reward or its features. In \cite{tai2018socially} and \cite{gupta2018social}, Generative Adversarial Networks were used for direct policy learning from demonstrations. These approaches provided advanced solutions to overcome the limitations mentioned above. However, these models required too much data for training \citep{che2020efficient}. On the other hand, learning from a small data set and generalizing to new configurations are desirable.

We also observe that most of the studies on the social navigation domain target only designing/learning the local controllers of the robots since they are responsible for producing motion commands. However, using only the local controller makes the robot vulnerable to the local minima problems \citep{koren}, and might fail to navigate the robot to its target position. Today, typical robotic navigation systems adopt the two-layered hierarchical approach for path planning tasks \citep{oreback2003evaluation}. A robot first calculates a trajectory in the so-called \textit{global planning} phase given an environment. Then, the robot follows the computed trajectory with a controller in the so-called \textit{local planning} phase.

This paper proposes a novel approach built on top of state-of-the-art neural network architecture, namely Conditional Neural Processes (CNPs) \citep{pmlr-v80-garnelo18a}. Given multiple demonstrations of a task, CNPs can encode complex multi-modal trajectories. CNPs extract prior knowledge directly from training data by sampling observations and predicting a conditional distribution over any other target points. Taking advantage of these capabilities, we extended CNPs in two dimensions: to generate complete navigation trajectories in the global planning phase and to generate goal-directed behaviors while actively avoiding pedestrians in the local planning phase. At both levels, our approaches produce trajectories that show the characteristics of the demonstrated ones. They can learn complex, nonlinear, and temporal relationships associated with external parameters and goals. Like other neural network-based learning systems, our system may fail to generate trajectories or control signals when it faces very different situations from the experienced ones, i.e., when it is required to extrapolate to novel situations outside the training range. To detect and react to conditions that may lead to failure, we propose continuously monitoring the environment using a failure prediction system, detecting situations outside the training range, and falling back to a hand-crafted reactive controller in case extrapolation is detected. We verified our system in a simulated mobile robot in different environments with static and moving pedestrians.

In the rest of this paper, we first give a literature review explaining the concepts used throughout this study. Then, we introduce our architecture in detail and elaborate on each system module. Later, we present our experiments and results. Finally, we conclude with a summary and future directions.

\section{Related Work}

\subsection{Hybrid Path Planning}
Traditionally, approaches to solving the path planning problem can be divided into two categories based on the environmental knowledge used: deliberate and reactive planning \citep{oreback2003evaluation}. Deliberate planners use environmental knowledge through static maps and compute the robot's trajectory before execution. On the other hand, reactive planners rely on sensory information to deal with obstacles in a local frame around the robot. Both approaches have advantages and disadvantages. Affected by the hybrid deliberate/reactive paradigm \citep{dudek2010computational}, a hybrid controller combining the two approaches has become a well-established approach to solving the path planning problem in recent years, \citep{murphy2019introduction}. In the following, we provide an overview of the two important building blocks of typical hybrid path planners, global and local planners, and the concept of social navigation.

\subsubsection{Global Path Planning}
In the first phase of standard hierarchical path planning, a global planning procedure is applied. On the static map of the environment, the task of a global planner is to create a path from the starting position to the destination. Global planners use utility functions to assign costs to the possible navigation trajectories they find. The use of utility functions allows the global planner to choose the trajectory with the desired properties, such as optimal length or time.

Prior to social navigation improvements, utility functions overlook the social aspects of the navigation trajectories of robots. On the other hand, the trajectories with optimal physical properties may not be preferred from a social point of view. The utility function of a more socially competent global planner optimizes the trajectories with respect to the social norms that humans follow \citep{kruse2013human}.

Conventionally, many graph search algorithms have been applied to compute the trajectory between initial and target configurations, the most popular being A* explained in \cite{astar}. For a complete list of global planning approaches, see \cite{giesbrecht2004global}. Global planning itself is not sufficient to navigate the robot between two points. Local planning is required to create velocity commands appropriate for the case of new or dynamic obstacles.

\subsubsection{Local Path Planning}
The local planning procedures are used in the second phase of hierarchical path planning to realize the computed trajectories. The main objective of the local planner is to generate velocity commands to allow the robot to move between the checkpoints of the precomputed trajectory. In addition, it is the task of the local planner to avoid the obstacles near the robot using sensory information about the environment of the robot. Avoiding obstacles requires a reactive control paradigm since it is impossible to consciously plan for dynamic obstacles such as humans or other robots. There are many local planning algorithms in the literature, such as \cite{vfh, dwa, apf, teb, zhu2006robot, vadakkepat2000evolutionary}. For a complete list, see \cite{cai2020mobile}.

Many traditional local planners are well suited for this task from a safety point of view. On the other hand, the traditional controllers do not consider social norms despite providing physical safety. They view people as obstacles to be avoided. Even though creating social plans at the global planning level increases the social aspect of navigation, it is essential to implement social maneuvers when the robot encounters a pedestrian. Recent attempts to create local planners that take these norms into account paved the way for more socially compliant local controllers \citep{ferrer2013robot, vasquez2014inverse, kretzschmar2016socially, kim2016socially}.

\subsection{Social Navigation}
\begin{figure}[t]
  \begin{center}
    \begin{subfigure}{0.49\columnwidth}
        \includegraphics[width=\columnwidth]{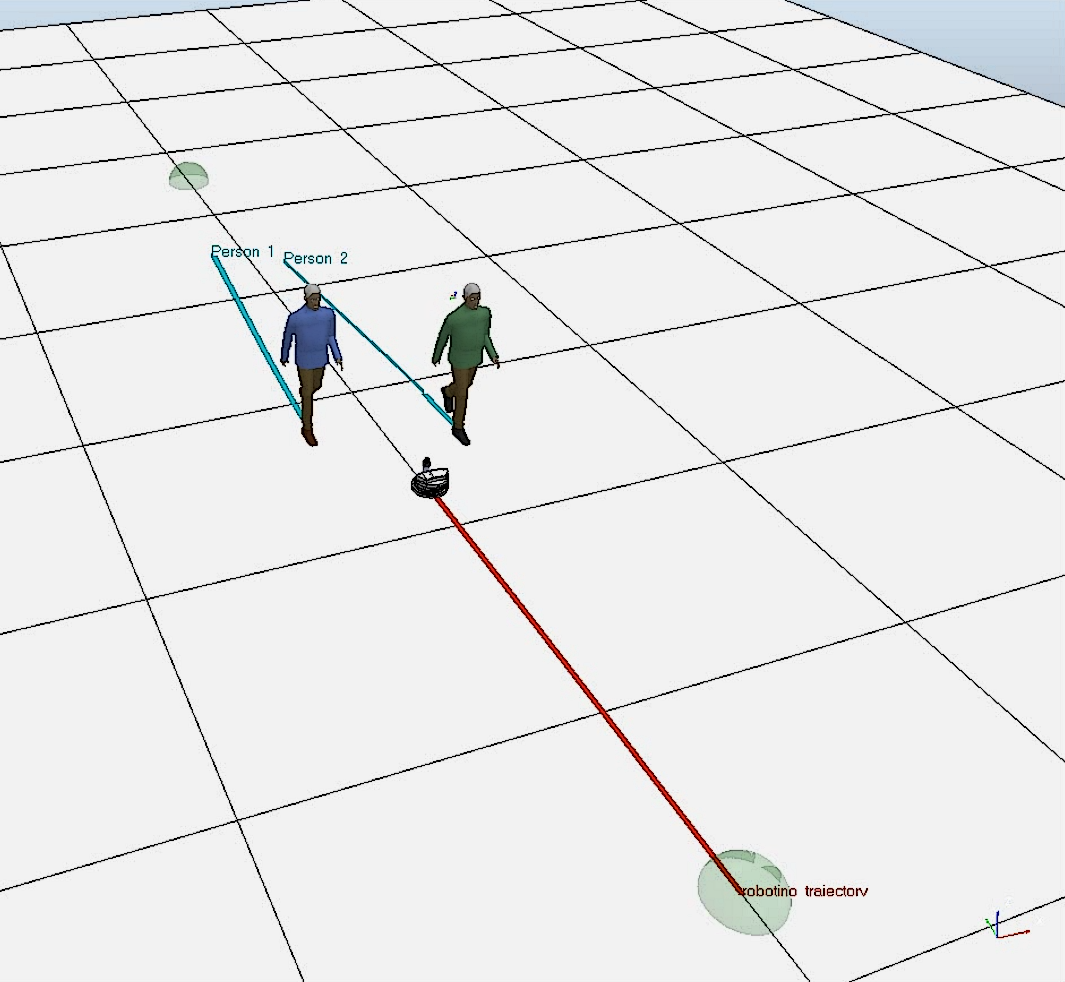}
    \end{subfigure}
    \hfill
    \begin{subfigure}{0.49\columnwidth}
        \includegraphics[width=\columnwidth]{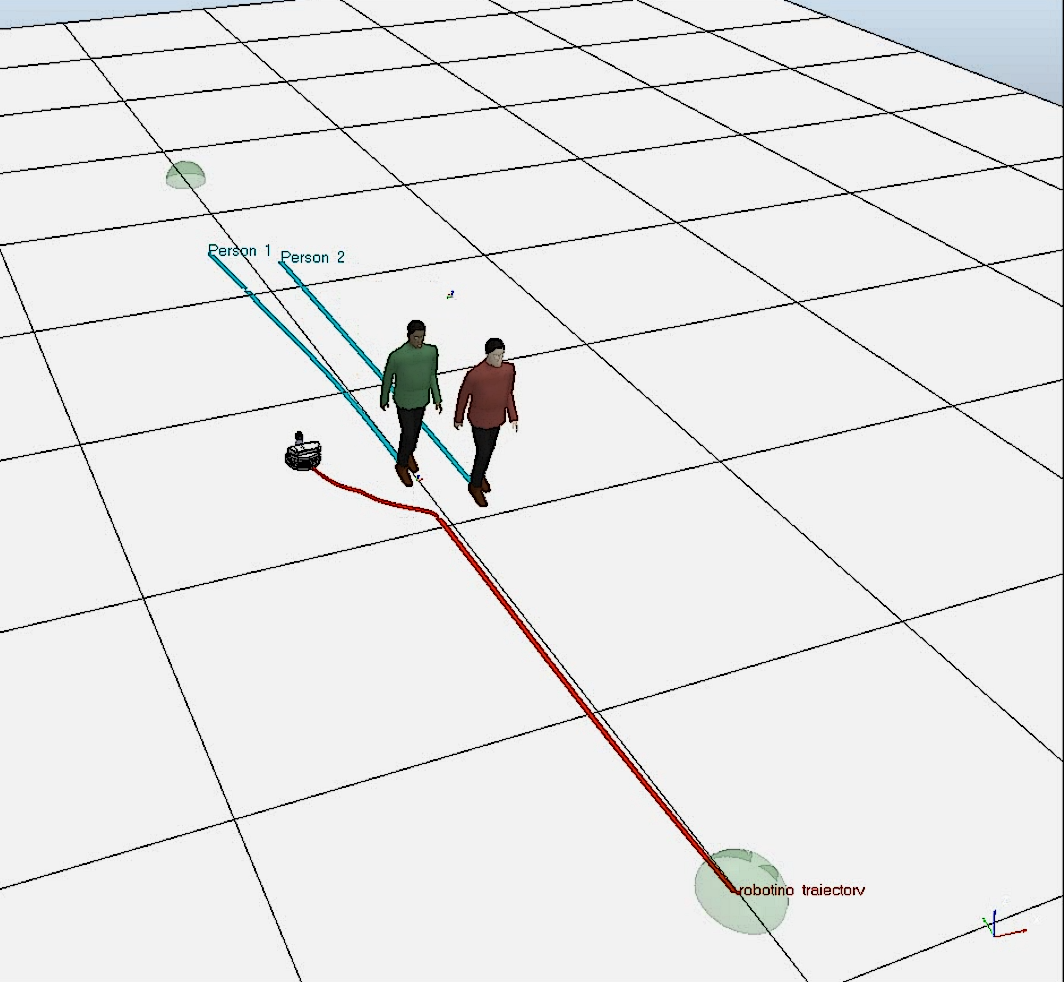}
    \end{subfigure}
    \vfill
    \begin{subfigure}{0.49\columnwidth}
        \includegraphics[width=\columnwidth]{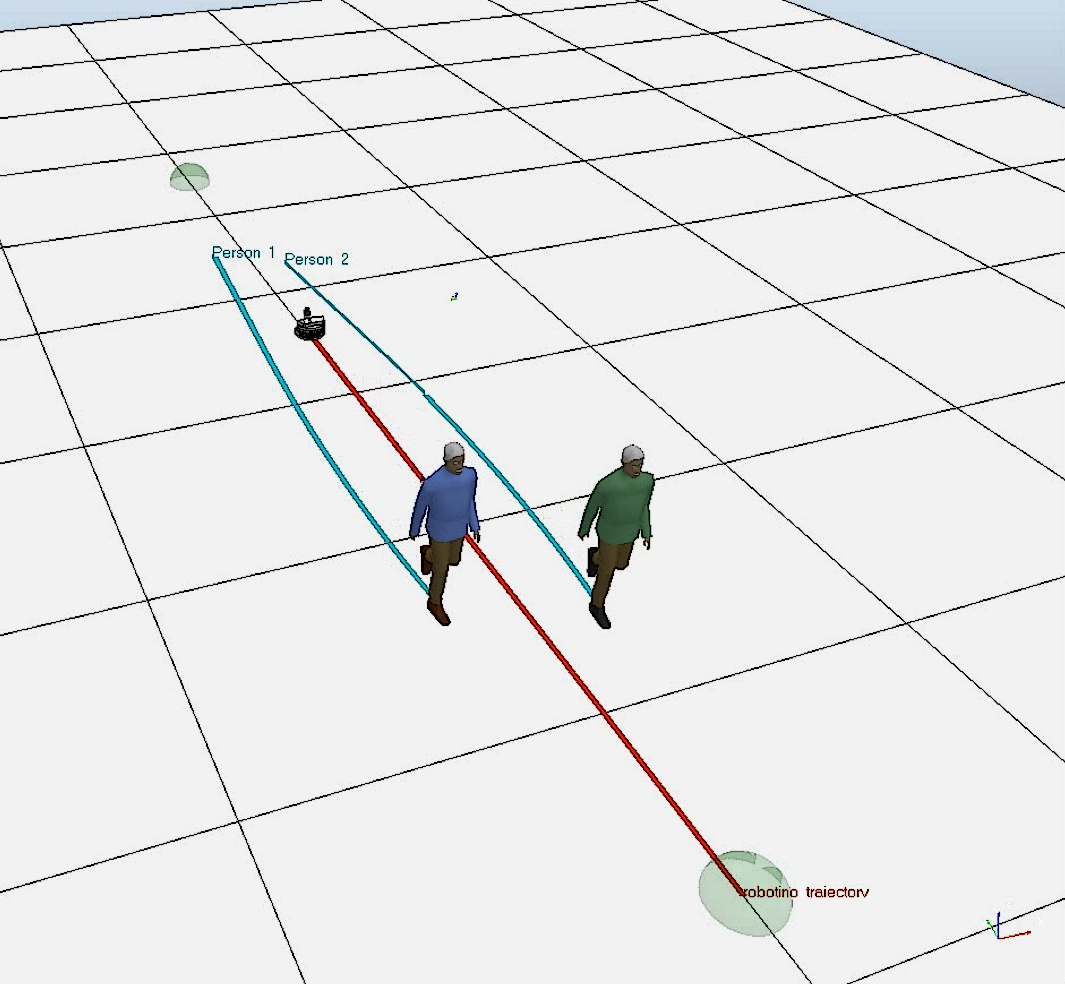}
    \end{subfigure}
    \hfill
    \begin{subfigure}{0.49\columnwidth}
        \includegraphics[width=\columnwidth]{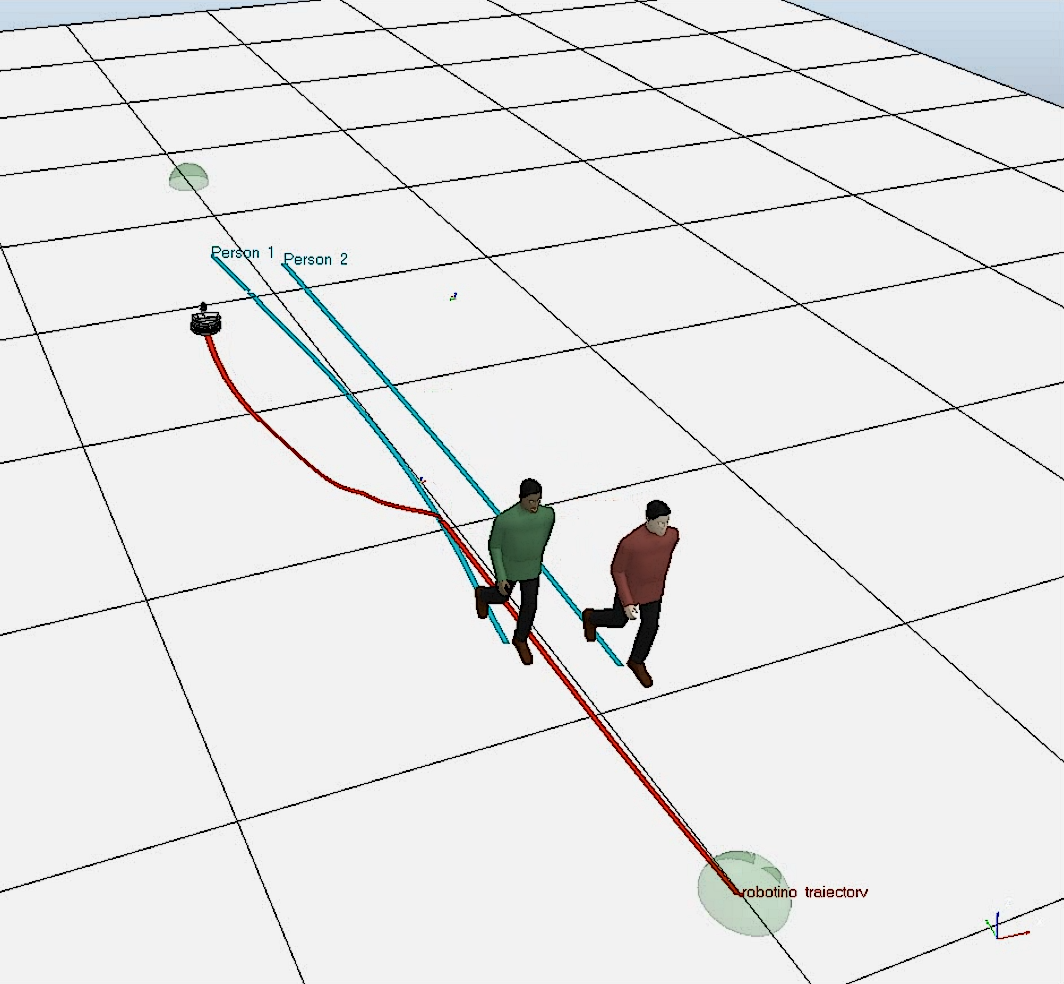}
    \end{subfigure}
    \caption{Comparison between the regular and social navigation. On the left, the robot passes between a group of two pedestrians, taking an energy-efficient trajectory. This behavior has the disadvantage of disturbing the people encountered on the way. The navigation trajectory on the right prioritizes the people's comfort instead of efficiency. Therefore, it is less disturbing and expected from socially compliant robots.}
    \label{fig:sn}
  \end{center}
\end{figure}

Social Navigation implies the exhibition of socially competent behaviors during the navigation of the robot. The nonverbal interaction caused by the navigation of the robot may be improved by the integration of social and cultural norms that people follow. \cite{kruse2013human} describe the benefits of social navigation as follows: it increases the comfort of the people around the robot, improves the naturalness of the robotic platform, and enhances the sociability of the robot. The concept of social navigation lies in the intersection of two fields: navigation and human-robot interaction. Figure \ref{fig:sn} presents a comparison of the purely safe approach with a socially compliant version. A robot with a perfectly safe navigation plan might disregard the importance of the comfort level of the encountered pedestrians, such as the one on the left. In contrast, although non-optimal, the executed navigation trajectory on the right is more socially compliant since it cares about the comfort level of the people around.

The early works in the domain are influenced by studies in social sciences. The concept of proxemics, introduced in \cite{hall1966hidden}, defines abstract social zones around the people and provides a basis for many studies in socially-compliant robot navigation \citep{syrdal2007personalized, huang2010human, lam2010human, asghari2010autonomous, mead2011proxemic}. Although these pioneer studies relied on the predefined set of rules to achieve social navigation, they drew attention to the subject and made it more popular.

\begin{figure*}
\begin{center}
\includegraphics[width=.7\textwidth]{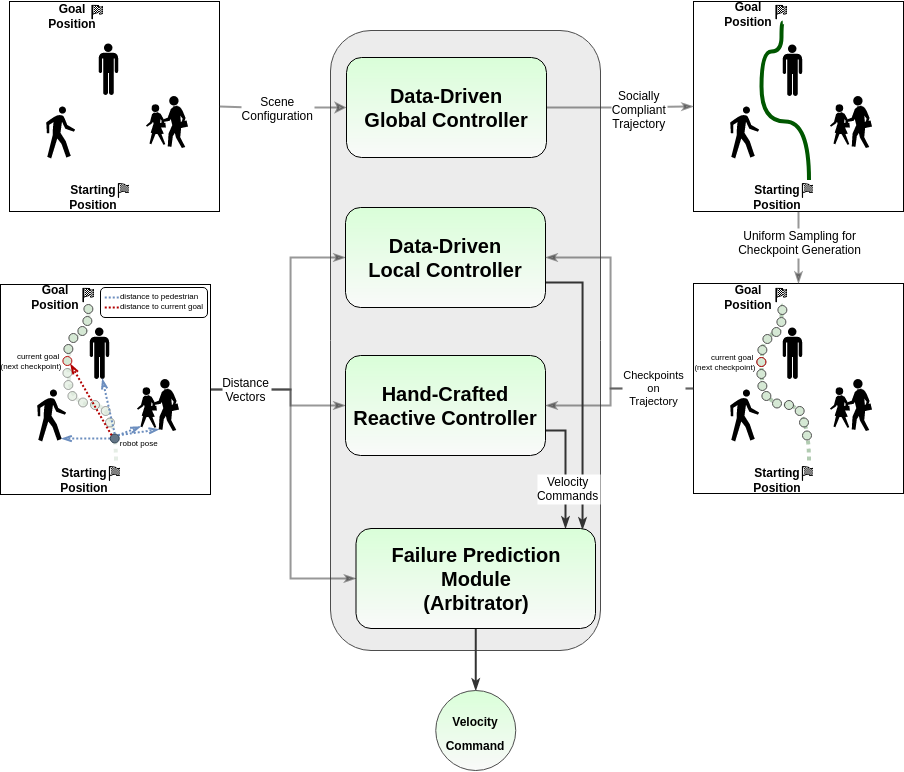}
\caption{The overview of the proposed navigation pipeline. The Data-Driven Navigation System is composed of four modules: Data-Driven Global Controller, Data-Driven Local Controller, Failure Prediction, and Hand-Crafted Reactive Controller. When a navigation task is given to the robot, the Data-Driven Global Controller generates a continuous navigation trajectory. On this trajectory, via-points are created, which must be reached one by one by the Data-Driven Local Controller. These modules are data-driven and are prone to extrapolation errors. Such cases are detected by the Failure Prediction Module, which transfers the control of the mobile robot to the Hand-Crafted Reactive Controller temporarily.}
\label{fig:system_overview}
\end{center}
\end{figure*}

Another important study in social sciences is the \cite{helbing1995social} where researchers introduce the Social Force Model (SFM) to explain the navigational behaviors of humans. They define a set of functions to calculate the local movements of pedestrians to reach a global goal. The simplicity of the SFM model causes many researchers in robotics to adopt the approach to move the robots as humans do, \citep{zanlungo2011social, ferrer2013robot, farina2017walking}.

As of late, data-driven approaches have become more prevalent in explaining human behavior since they are more adaptable to many situations. Many researchers apply Inverse Reinforcement Learning (IRL) to calculate a reward function that describes the navigational behavior of the human, \citep{kitani2012activity,vasquez2014inverse,kim2016socially}. Another stream of work uses Deep Reinforcement Learning to generate human-aware robot navigation, \citep{chen2017}. These studies assume the availability of either the reward or the features that compose the reward. To relax this assumption, approaches that learn the social norms merely from the data became more popular.

In \cite{social_lstm}, researchers use networks with Long-Short Term Memory cells (LSTMs). Later, an improved version of this study is presented in \cite{gupta2018social}, where researchers use LSTMs inside a generative adversarial setting. These studies successfully predict human navigation trajectory in a limited local frame. Despite being considerably close to the social navigation domain, the trajectory prediction methods in these studies cannot be directly applied to mobile robots. A robot placed in a real-world environment has to meet serious time-complexity considerations. Moreover, these approaches are limited to local frames, while path-planning on mobile robots requires a global navigation goal \citep{latombe1991robot}.

Data-driven approaches generally process navigation trajectories in datasets to realize social navigation. However, as mentioned in \cite{mavrogiannis2021core}, the scarcity of established datasets has been a significant issue in the domain. Until recently, many studies in the literature (e.g. \cite{trautman2010unfreezing, vemula2018social}) have been using pedestrian datasets to train models, such as \cite{lerner2007crowds} and \cite{pellegrini2009you}. Although such datasets provide real-world interactions among pedestrians, they do not contain human-robot interaction. Therefore, pedestrian datasets may overlook any possible effect of an embodied robotic agent's presence on a neighboring pedestrian's navigation trajectory. Lately, we have witnessed an increase in datasets focusing on human-robot interactions for social navigation studies. \cite{manso2020socnav1} aims to evaluate the comfort levels of pedestrians around a robot in an environment. On the other hand, this dataset does not capture the navigation dynamics of the robot and pedestrians as it merely focuses on static scenes. Recent datasets, such as \cite{yz17iros}, \cite{martin2021jrdb}, \cite{karnan2022socially}, address both shortcomings, where researchers control robots by teleoperation in real-world environments to record navigation demonstrations.

\section{Our Method}

\begin{figure*}[t]
\begin{center}
\includegraphics[width=0.9\linewidth]{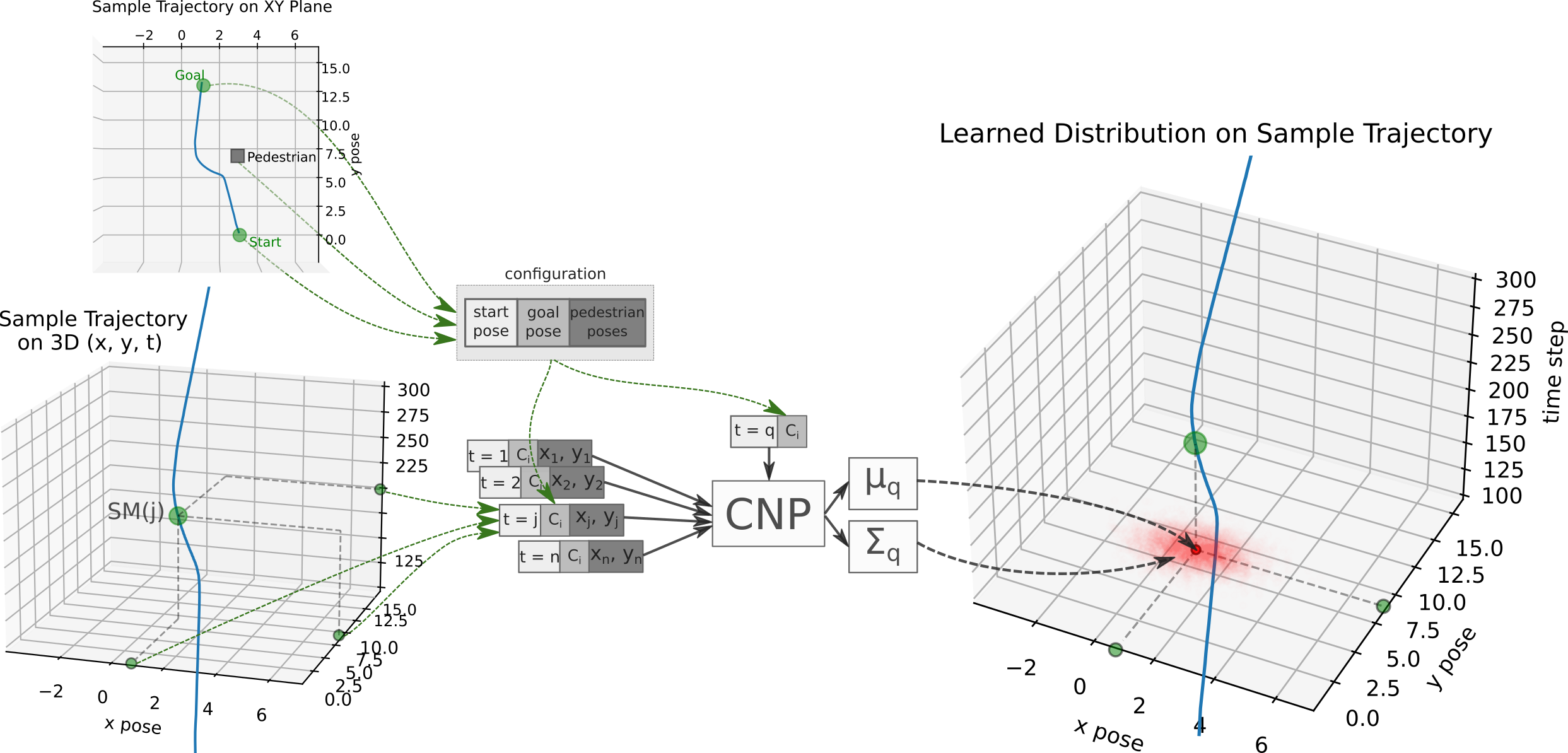}
\caption{Training the Data-Driven Global Controller with a demonstration trajectory. The configuration of the environment and the demonstration trajectory on this environment are gathered and processed. Later, this data is given as input to the CNP structure. The model predicts a bivariate normal distribution of 2D positions on the XY plane for a queried time step. The log-likelihood of the actual data point given the predicted distribution produces the loss that is back-propagated to tune the weights of the neural network structure.}

\label{fig:cnp_global_tr}
\end{center}
\end{figure*}

This work proposes a hybrid data-driven navigation system that uses advanced neural techniques in global and local layers coupled with a novel execution monitoring module for fault-tolerant intelligent navigation.
Figure \ref{fig:system_overview} demonstrates the overview of our proposed system. Our navigation system consists of four modules: Data-Driven Global Controller, Data-Driven Local Controller, Failure Prediction, and Hand-Crafted Reactive Controller. After introducing these modules in this section, the following subsections will provide the details. First of all, given a target position, the Data-Driven Global Controller is responsible for producing a complete navigation trajectory consistent with the demonstration trajectories previously provided to the robot. Next, several via-points are uniformly sampled from the generated navigation trajectory. The second module, namely the Data-Driven Local Controller, is responsible for generating motion commands to reach each via-point one by one. While maneuvering the robot, this controller reacts to the dynamic changes in the vicinity of the robot and changes the target to the next via-point as soon as the current target is reached. This procedure continues until the robot reaches the final target, i.e., the given goal position.  Suppose the local controller fails to follow the trajectory at any point; a pedestrian may be blocking the path, for example. In that case, the global controller can be called again to recalculate a new trajectory on the same navigation task. The first two modules use data-driven approaches, which learn from provided trajectories, and can be used to learn social competencies for social navigation. The proposed navigation system achieves learning and control in these modules using a neural network family. Similar to other neural network architectures, while models can interpolate to novel situations within the training range, the learned models are prone to extrapolation errors when they run on inputs outside their training range. If unattended, such errors can cause collisions with pedestrians during navigation. Two additional modules are incorporated into our system to detect and react to the extrapolation cases before any collision or failure happens: the Failure Prediction Module and the Hand-Crafted Reactive Controller. The Failure Prediction Module observes the entire operation of the system and is responsible for detecting outlier situations. If this module detects that data-driven controllers are queried with an input outside their training range, the control is temporarily transferred to the Hand-Crafted Reactive Controller, ensuring the safety of navigation. When the risk of failure vanishes, i.e., the robot and its environment are detected to be in the training range again, the data-driven modules take back the control of the robot. These components are combined to form a full-fledged navigation pipeline. In the following, we explain these modules in detail.

\begin{figure*}[t]
\begin{center}
\includegraphics[width=\textwidth]{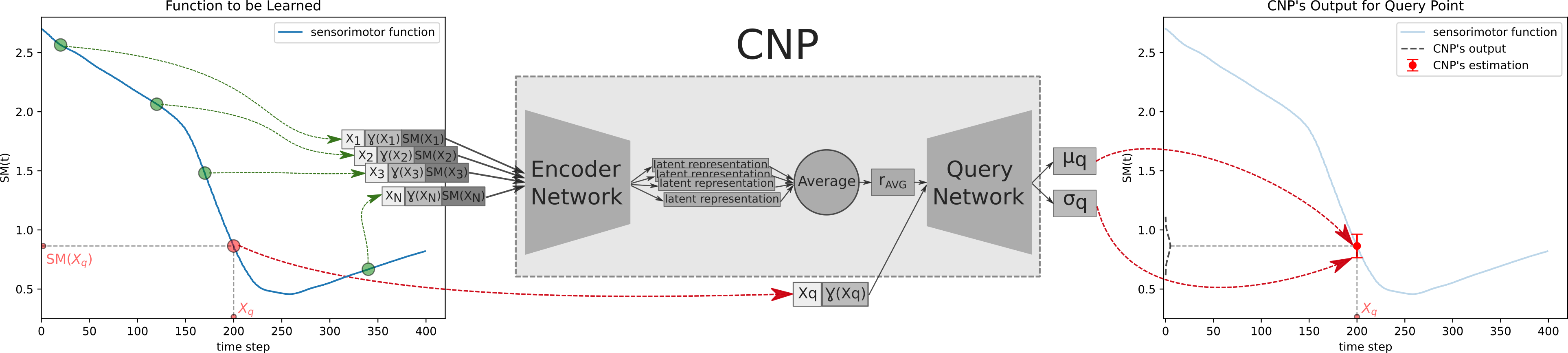}
\caption{The general layout of the training phase of our model. First, a random number of observation points are sampled randomly on $\textit{SM}()$ trajectory. An observation point consists of the input ($X$), the task parameter for that input ($\gamma(X)$), and the value of the function for that input ($\textit{SM}(X)$). Observation points are fed to the Encoder Network. Obtained latent representations are averaged to retrieve a compact representation of the trajectory. The query network is run with a random query point to produce the prediction of the system. This prediction causes a loss that is back-propagated through both networks.}
\label{fig:cnp_general}
\end{center}
\end{figure*}

\subsection{I - Data-Driven Global Controller}

In our work, the task of global planning is to generate a complete trajectory for navigation. A set of discrete via-points can be sampled from this trajectory to enable the local planner to reach each via-point successively. The Data-Driven Global Controller aims to learn a parametric distribution of social navigation trajectories that reflect the social norms of the people. Therefore, after learning the underlying distribution of socially acceptable trajectories, this module can reactively change the navigation trajectory according to the configuration of the environment.

\textbf{Training the Data-Driven Global Controller:} We recorded a set of expert demonstrations to teach the trajectory generation function to our model. Each demonstration contains an environment configuration and a navigation trajectory close enough to the optimal social behavior in the corresponding environment. The environment configuration is the concatenation of the following three elements: the robot's start and goal positions and the pedestrians' positions. In addition, navigation trajectories are stored as a set of \textit{(x, y)} coordinates with corresponding time steps.

% \remyigit{Formally, the set of all trajectories is a collection of \textit{N} expert demonstrations, denoted by $D = \{\tau^1, \tau^2, \dotsc, \tau^N\}$. Each navigation trajectory $\tau$ is a set of 2D positions at each time step; $\tau^i = \{(x_1^i, y_1^i), (x_2^i, y_2^i), \dotsc, (x_T^i, y_T^i)\}_{i=1}^N$. Furthermore, the configuration $C$ of each scene is formed as $C_i = \{(x_{start}^i, y_{start}^i), (x_{goal}^i, y_{goal}^i), [(x_{p_j}^i, y_{p_j}^i)]_{j=1}^{P}\}_{i=1}^N$, where $(x_{p_j}, y_{p_j})$ represents the 2D position of one of the $P$ pedestrians in the scene.
% }

Formally, the set of all trajectories is a collection of \textit{N} expert demonstrations, denoted by $D = \{\tau^i\}^N_{i=1}$. Each navigation trajectory $\tau$ is a set of 2D positions at each time step; $\tau = \{(x_j, y_j)\}_{j=1}^T$. Furthermore, the configuration $C$ of each scene is formed as $C = \{(x_{start}, y_{start}), (x_{goal}, y_{goal}), ((x_{p_1}, y_{p_1}) \dotsc (x_{p_K}, y_{p_K}))\}$, where $(x_{p_k}, y_{p_k})$ represents the 2D position of one of the $K$ pedestrians in the scene.

The overall training procedure is depicted in Figure \ref{fig:cnp_global_tr}. At training time, the system uniformly samples a navigation trajectory $\tau^i$ from $D$. Afterward, $n$ observation points are uniformly sampled on this trajectory, where $n$ is a random number between 0 and $n_{max}$. The navigation trajectory is the sensorimotor function that the model learns in our representation. It is a function of time and is shaped by the scene's configuration. The realizations of this function on different time steps correspond to 2D positions in the real world. That is, $\tau = \{\textit{SM}(t)\}_{t=1}^T$, where $\textit{SM}(t) = (x_t, y_t)$. Moreover, we represent the configuration, $C$, of the environment with the task parameter, using the model's $\gamma$ function. Hence, in this representation, $\gamma(t) = C$.

Figure \ref{fig:cnp_general} introduces the underlying neural network model with the corresponding Encoder-Decoder structure. The Encoder Network takes in $(t, \gamma(t)$ and $\textit{SM}(t))$ tuples, produces their corresponding latent representations, and applies averaging operation to generate a general representation to encode the n tuples. On the other hand, the Query Network is responsible for generating the distribution related to the position of the robot (\textit{SM}($t_q$)) at any time point $t'$ where the generated position is required to be consistent with the average latent representation of n tuples. Therefore, given the average latent representation of n $(t, \gamma(t)$ and $\textit{SM}(t))$ tuples and a random target time point $t_q$, the Query Network produces the distribution of the predicted position of the robot, i.e., outputs a bivariate normal distribution with parameters $(\mu_q, \Sigma_q)$. With this output, the complete neural network model is trained using the following loss function:

\begin{equation}
\label{eq:loss}
\mathcal{L} = -\: log\: P(\:\textit{SM}(t_q)\: |\: \mu_q,\: softmax(\Sigma_q)\:)
\end{equation}
where SM($t_q$) corresponds to the actual (x, y) coordinates of the trajectory at time $t_q$.

\begin{figure}
\begin{center}
\includegraphics[width=\columnwidth]{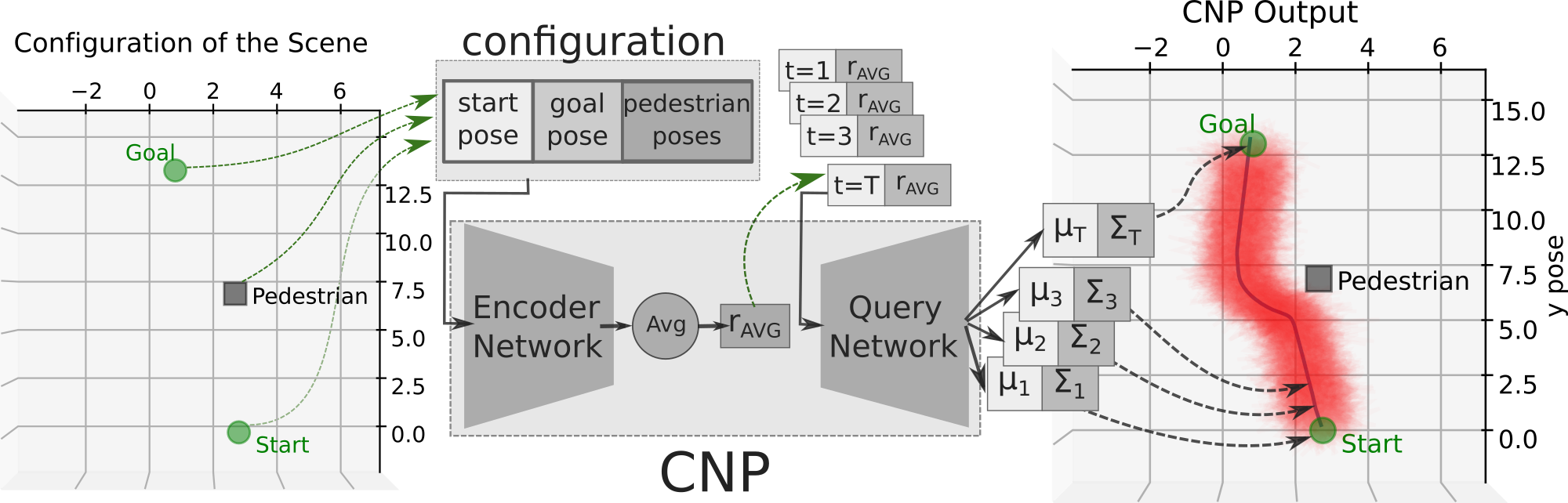}
\caption{Generating an entire navigation trajectory for the configuration given on the left panel. This configuration is given as input to the network, as shown in the middle column. The predicted trajectory is shown on the right. Note that the solid line corresponds to the sequence of the mean positions and the red shaded area illustrates the variance information. For details, please refer to the text.}
\label{fig:cnp_global_test}
\end{center}
\end{figure}

\textbf{Querying the Trained Data-Driven Global Controller:} After the training of the Data-Driven Global Controller with given navigation trajectories, it can be queried to generate new navigation trajectories given new environment configurations, as illustrated in Figure \ref{fig:cnp_global_test}. The underlying system can predict the position of the robot at any time point. Therefore, the time step is used as the input, while the scene configuration is used as the task parameter to generate the corresponding position of the desired trajectory. The Data-Driven Global Controller can be queried simultaneously and independently for all time points. In the end, this module outputs an entire trajectory of 2D coordinates - from the starting position to the goal position- as a function of time. When trained with social navigation trajectories, the learned function is a social trajectory generator that avoids pedestrians in the intended course of the navigation.

\subsection{II - Data-Driven Local Controller}

Typically, the main task of a local planner is to move the robot through the via-points on the navigation trajectory computed by the global planner. Moreover, it is the task of the local planner to navigate the robot without colliding with pedestrians. In our system, the Data-Driven Local Controller is used to accomplish both tasks and preserve the characteristics present in the demonstrations.

This module is built on top of a CNP-based neural architecture. As in traditional local planners, this module generates goal-directed behavior by targeting short-distance via-points as intermediate goals. It uses the relative position of the next checkpoint as input, $X$. To achieve collision-free navigation, relative positions of pedestrians are passed to the network as a task parameter, $\gamma(X)$. This gives the Data-Driven Local Controller the ability to reactively adjust its output with respect to the changing pedestrian positions.

In addition to the basic tasks, the neural network within the Data-Driven Local Controller can learn the social characteristics present in the local evasive maneuvers of the people, using only the data from their navigation trajectories. In the end, the Data-Driven Local Controller decides which action to take in the form of velocity commands of the mobile base.

\begin{figure}[t]
\begin{center}
\includegraphics[width=\columnwidth]{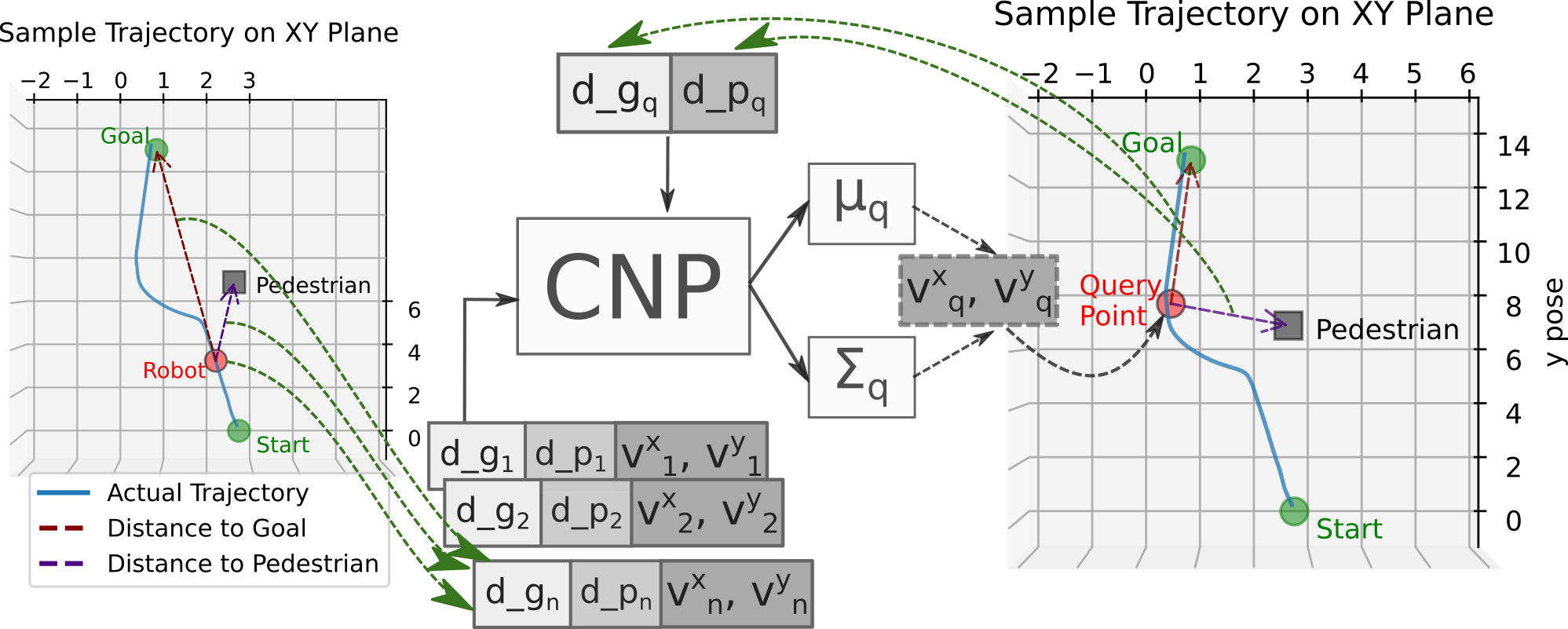}
\caption{Training the Data-Driven Local Controller with a demonstration trajectory. Along a demonstration trajectory, several observations are sampled randomly. Each observation contains the relative position of the goal, $d\_g$, relative positions of the pedestrians, $d\_p$, and the velocity command, $v$. The velocity prediction of the model for the query point provides the loss, which is back-propagated through both networks of the CNP structure.}
\label{fig:cnp_local_tr}
\end{center}
\end{figure}

\textbf{Training the Data-Driven Local Controller:} The local planners use sensory information to understand their local frames. Our system processes the low-level sensory information to obtain high-level parameters, such as relative position of pedestrians and relative position of the goal position. The Data-Driven Local Controller reactively responds to the changes in these high-level parameters. More importantly, it does so while preserving the characteristics present in the previously learned demonstrations. Therefore, when the module is trained with socially acceptable trajectories, it can learn social norms. The demonstration trajectories are recorded as a set of tuples ($d_{goal}$, [$d_{pedestrian}$], $v$), where $d_{goal}$ is a 2D vector of the relative position of the current goal, [$d_{pedestrian}$] is an array of 2D vectors representing relative position of the pedestrians in the scene, and $v$ is a 2D vector representing the velocity of the base on the 2D plane.

Prior to the training, trajectories are processed and converted into a set of observations that the neural network of the Data-Driven Local Controller uses. An observation is formed by the concatenation of three parts: $X, \gamma(X)$ and $\textit{SM}(X, \gamma(X))$, where $X = d_{goal}$, $\gamma(X) = [d_{pedestrian}]$, and $\textit{SM}(X, \gamma(X)) = (v^x, v^y)$, and $v^x, v^y$ correspond to X and Y components of the velocity vector. Figure \ref{fig:cnp_local_tr} illustrates the training procedure. The training process uniformly samples a trajectory $\tau^i$ from the set of demonstrations, $D$. $n$ observations are uniformly sampled, where $n$ is a random number between 0 and $n_{max}$. These observations are fed into the \textit{Encoder Network} to obtain latent representations, which, in turn, are averaged to create the average latent representation of the entire trajectory. This is given as input to the \textit{Query Network}, along with a random query point $X_q$, and the value of the task parameter for this query point, $\gamma(X_q)$. The \textit{Query Network} outputs a bivariate normal distribution with parameters $(\mu_q, \Sigma_q)$ which represents the normal distribution that describes the model's prediction about the velocity command. The loss calculation is the same as in Equation \ref{eq:loss}, which is back-propagated through the entire CNP structure.

Note that in the previously described Data-Driven Global Controller, latent representation encoded the entire trajectory for the corresponding environment and the navigation task and was conditioned with successive time points (independently) to generate the successive points of the trajectory that the robot is supposed to follow. In the Data-Driven Local Controller, on the other hand, the latent representation encodes instantaneous control command rather than the entire trajectory. It, therefore, is conditioned on the relative position of the checkpoint to generate the corresponding velocity command.

\textbf{Querying the Trained Data-Driven Local Controller:} After training the encoder and the query networks, the model can be run for specific scene configurations. Conditioned on the start and goal positions, the use of the task parameter $\gamma(X)$ gives the model the ability to reactively change the velocity commands with respect to changing pedestrian positions and possibly changing checkpoint positions in case of the re-computation of the trajectory. At any time, the relative position of the robot with respect to the current goal ($d_{goal}$) and the closest pedestrian ($d_{pedestrian}$) can be concatenated and given to the model as input. The neural network outputs the velocity command $(v^x$, $v^y)$ to be executed by the robot. At test time, the query is fast enough to be used in real-time applications. Therefore, this module can be used as a local controller that generates velocity commands following the characteristics of the demonstrations, which is used for social navigation.

\subsection{III - Failure Prediction Module}

The previously described Data-Driven Global and Local Controllers learn representations inside the domain they are trained. Therefore, these modules are inherently prone to extrapolation cases when queried outside the learning range. These cases may lead to anomalous behavior and, in practice, generates undesirable velocity commands. Although infrequent, such anomalies may cause collisions and are unacceptable in navigation systems where the safety of pedestrians is of the utmost importance.

To avoid such undesired behavior, we provide the navigation system with the capability to continuously observe the states it encounters during a navigation task for their likelihood of lying outside its training domain. Firstly, we leverage the ability of Generative Adversarial Networks (GAN) \citep{goodfellow2014generative} to learn the input-space distribution. There are many variants of GANs in the literature, such as Wasserstein GAN (W-GAN) \citep{arjovsky2017wasserstein}. For our case, standard GAN worked more stably compared to the W-GAN.

\begin{figure}[t]
  \begin{center}
    \begin{subfigure}{\linewidth}
        \includegraphics[width=\linewidth]{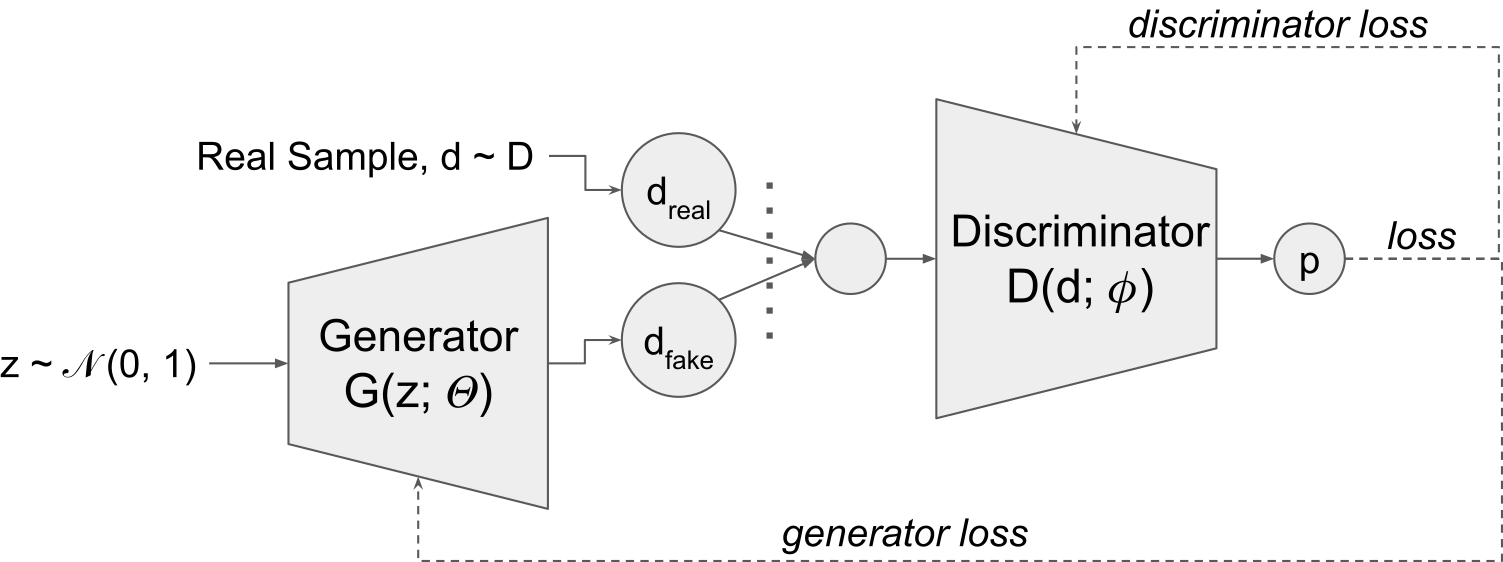}
        \caption{GAN training for predicting the extrapolation. While the actual data is sampled from demonstration trajectories, the generator network produces realistic candidates. The discriminator network learns the distribution that the navigation data belongs. As a result, it learns to attribute small probabilities to cases with unusual pedestrian positions.}
        \label{fig:fpma}
    \end{subfigure}
        \begin{subfigure}{\linewidth}
        \includegraphics[width=0.85\linewidth]{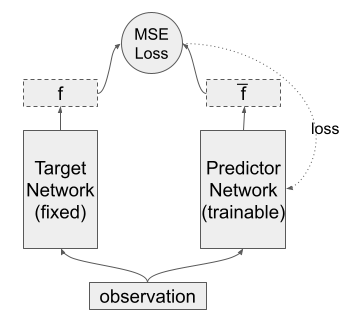}
        \caption{RND training for predicting the extrapolation. The observation is sampled from demonstration trajectories. The target network outputs random values, and the predictor network learns to mimic them. After training, two networks output similar values for familiar observations.}
        \label{fig:fpmb}
    \end{subfigure}
    \caption{Approaches used in the Failure Prediction Module. The RND approach is preferred in this study due to its stability.}
  \end{center}
\end{figure}

Figure \ref{fig:fpma} depicts the training procedure. On the same dataset that the local CNP is trained, we train a GAN. GAN takes as input [$d_{pedestrian}$] from actual trajectories from the dataset. The generator network attempts to produce realistic candidates to deceive the discriminator network in deciding whether its input belongs to the actual dataset or not. The objective function, given in Equation \ref{eq:gan_loss}, follows the standard minimax loss introduced in \cite{goodfellow2014generative}.

\begin{equation}
\label{eq:gan_loss}
\underset{G}{\text{min}}\; \underset{D}{\text{max}}\; \mathbb{E}_{d}\left[\log D(d)\right] + \mathbb{E}_{z}\left[\log \left(1-D(G(z))\right)\right]
\end{equation}

During the training of the networks with their corresponding losses, the generator network becomes better at producing realistic candidates, and the discriminator network becomes better at discriminating against them. After the training, the discriminator learns the distribution that the actual data belongs to. We use the output of the discriminator network, which is the probability of an encountered state being sampled from the same distribution as the training data, in predicting the extrapolation.

Although the system benefits from using GANs, these architectures are often criticized for being unstable due to oscillation and mode collapse, \cite{goodfellow2016nips}. Therefore, we have implemented another approach, named Random Network Distillation (RND) \citep{burda2018exploration} in this module for detecting out-of-distribution (OOD) samples. This approach uses two sub-networks; the first one - the target network - is a fixed and randomly-initialized multi-layer perceptron (MLP) and the second - the predictor network - is a standard fully connected MLP. The approach evaluates the novelty of an encountered system by calculating the distance of the output that two sub-networks produce. Initially, the two would produce different results. By training the predictor, it starts to output similar values as the target network does. Upon training, when an OOD input is queried, two sub-networks output very different values, producing a high error value and enabling our system to predict extrapolation errors before they occur. The structure of this architecture is given in Figure \ref{fig:fpmb}.

The model is trained with the relative positions of pedestrians and the goal at each time step. This information is extracted from the same expert demonstrations that the data-driven controller is trained with. After training the RND model, the distance between its sub-networks is used to detect whether a new data point is from the training range.

\subsection{IV - Hand-Crafted Reactive Controller}
Whenever the Failure Prediction Module outputs a small probability for an observation point, the navigation system concludes that it is about to extrapolate. This usually leads to erratic movements and potentially a collision with a pedestrian. To control the robot safely in those situations, we created the Hand-Crafted Reactive Controller. This module implements the Social Force Model (SFM) to move the robot in a safe and socially compliant manner. On the other hand, the SFM is not flexible enough as many real-world applications require since the attractive and repulsive components are defined and tuned manually. Therefore, the pipeline uses the Hand-Crafted Reactive Controller as a fallback module. During the period it controls the robot, the Failure Prediction Module continues to check whether it is safe for the Data-Driven Local Controller to take back control.

\begin{figure}[b]
\begin{center}
\includegraphics[width=0.9\columnwidth]{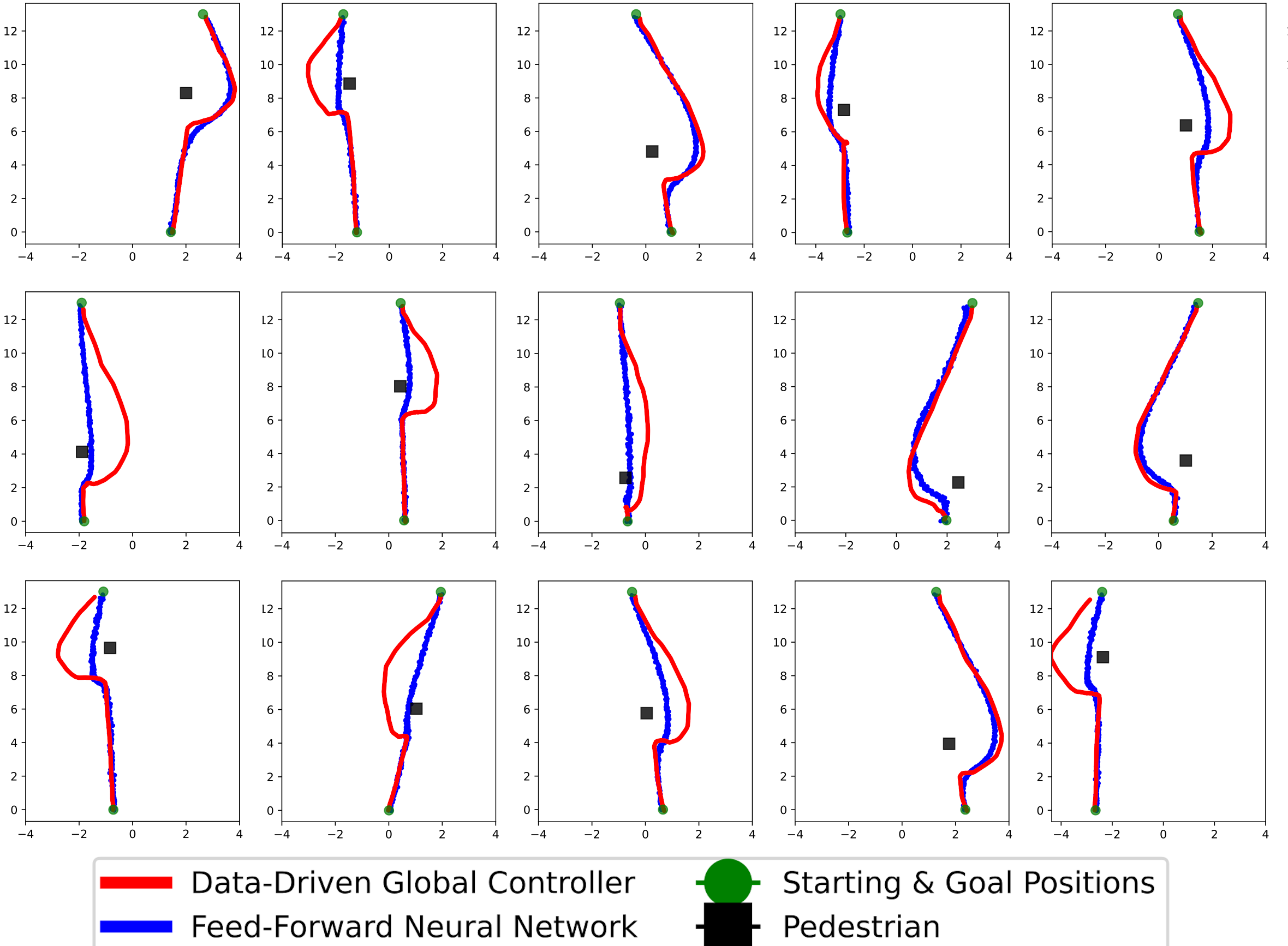}
\caption{The comparison of trajectories generated by the Data-Driven Global Controller and a feed-forward neural network on several environment configurations. Results show that the Data-Driven Global Controller produces less disturbing trajectories for pedestrians.}
\label{fig:cnmp_nn}
\end{center}
\end{figure}

\section{Experiments and Results}
Our system was verified in the CoppeliaSim simulation environment \citep{coppeliaSim} that includes an omnidirectional robot platform, namely Robotino \citep{robotino}. For details about the dataset used in training all deep models, including CNPs, GAN and RND, please refer to Appendix \ref{app:dat}. Also, hyperparameters used in training all networks are given in \ref{app:mod}.

\subsection{Analysis of the Generated Global Trajectories}

The application of neural networks in global planning tasks is an established procedure. Many studies in the literature propose variants of neural networks that optimize for the shortest or the minimum-energy trajectories \citep{glasius1995neural}. In this part, we aim to emphasize the capability of our structure over the standard neural network approach from the perspective of social navigation.

On the dataset of trajectories described in Appendix \ref{app:dat}, we trained our Data-Driven Global Controller and a standard feed-forward neural network (FFNN) with five layers for comparison. After training, both networks were queried in novel environments with different starting and goal positions and with different pedestrian positions. A subset of the trajectories generated by both networks is given in Figure \ref{fig:cnmp_nn}. In most cases, both approaches could generate trajectories that allow the robot to avoid pedestrians, which are shown with black dots. In many cases, though, the FFNN failed to generate global paths that avoid the pedestrian in a socially compliant manner, whereas our system could. We believe that the difference in the performance was due to the inability of standard FFNNs to encode multiple modes of operations, in other words, multiple trajectories for the same or very similar environments. In detail, given multiple trajectories that avoid the same pedestrian from different sides, our system can encode different modes in its robust latent space. In contrast, standard FFNNs learn an average representation from multiple trajectories in the same environment. Therefore, the produced trajectory by FFNNs corresponded to the average trajectory of the demonstrated ones.

\begin{figure}[t]
\begin{center}
\includegraphics[width=0.6\columnwidth]{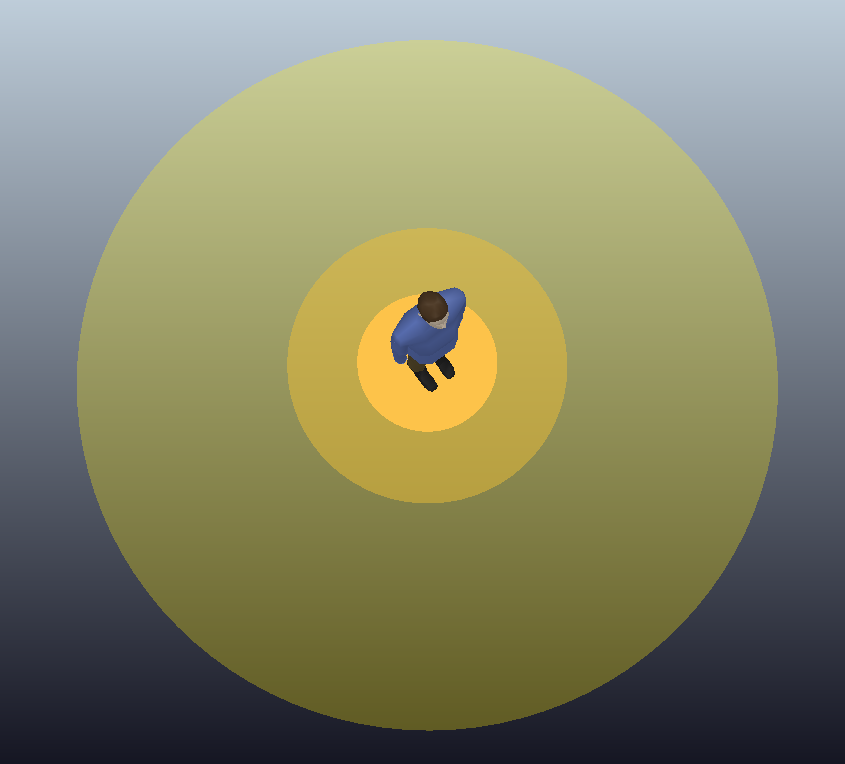}
\caption{Illustration of the proxemics zones. The theory of proxemics defines four notional zones surrounding a human that influence the behavior of the individual by affecting the type and the content of the interaction. These zones (from inner to outer) are Intimate (0-0.5 m.), Personal (0.5-1 m.), Social (1-4 m.), and Public (4-8 m).}
\label{fig:prx}
\end{center}
\end{figure}

In order to further analyze and compare these approaches from the social navigation perspective, we used the metrical comparison on the concept of proxemics. Defined by \cite{hall1966hidden}, the theory of proxemics investigates the spatial aspects of nonverbal communication. It suggests that interpersonal spaces are critical determinants of social interactions. In addition, there are four abstract zones surrounding a human. An illustration of them is given in Figure \ref{fig:prx}. The physical intrusion of these zones is only allowed under certain circumstances; for example, people can comfortably allow their family and close friends in their personal zones. On the contrary, disallowed intrusion of these zones usually leads to discomfort or anxiety.

\begin{figure}[h]
  \begin{center}
    \begin{subfigure}{0.75\linewidth}
        \includegraphics[width=0.75\linewidth]{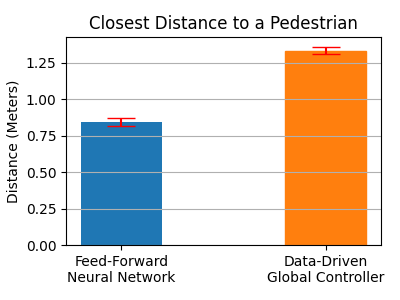}
        \caption{}
        \label{fig:meta}
    \end{subfigure}
        \begin{subfigure}{0.75\linewidth}
        \includegraphics[width=0.75\linewidth]{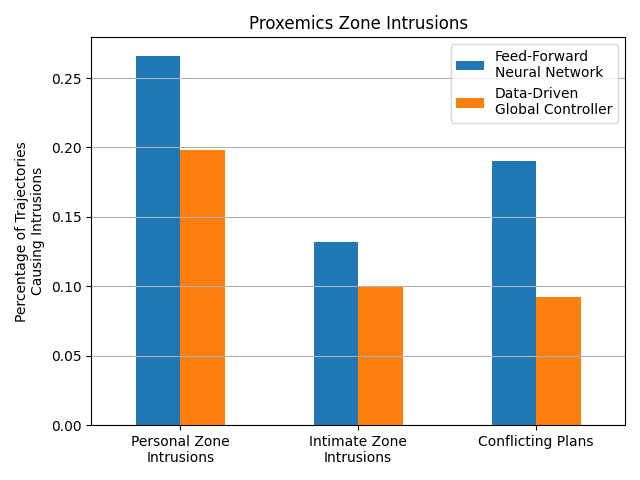}
        \caption{}
        \label{fig:metb}
    \end{subfigure}
    \begin{subfigure}{0.75\linewidth}
        \includegraphics[width=0.75\linewidth]{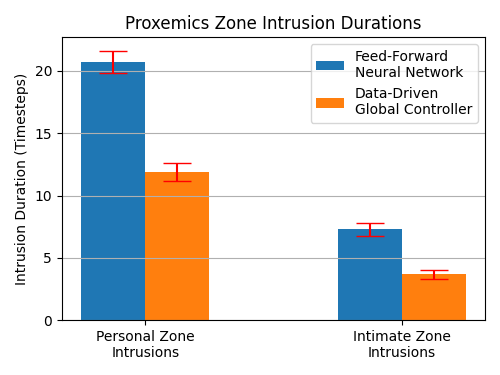}
        \caption{}
        \label{fig:metc}
    \end{subfigure}

    \caption{Comparison of approaches in terms of distance-based metrics. Figure \ref{fig:meta} shows the closest distances to a pedestrian both approaches produce. Figures \ref{fig:metb}, \ref{fig:metc} show counts and durations of zone intrusions. Trajectories generated by the Data-Driven Global Controller pass close to the pedestrian less frequently, causing less discomfort. Proxemics-zone intrusions support this argument. Refer to the text for more details.}
    \label{fig:metrics}
  \end{center}
\end{figure}

We present the metrical comparison of these approaches in Figure \ref{fig:metrics}. Using 1000 randomly generated environments, we first measure the closest distance of the generated trajectories to a pedestrian in the environment and compare the methods based on the closest distances. Figure \ref{fig:meta} shows the comparison. The Data-Driven Global Controller is better at producing trajectories that do not come too close to pedestrians. The closer the distance to a pedestrian, the more disturbing the robot becomes \citep{diego2011please}.

Moreover, \cite{vasquez2014inverse} uses a distance-based comfort metric to compare navigation approaches from the social navigation perspective: the number of intrusions of the intimate and personal zones. In addition to that, we also use the duration of zone intrusions as an indicator of discomfort. A more extended period of intrusion of proxemics zones naturally leads to a higher level of anxiety. As shown in Figures \ref{fig:metb} and \ref{fig:metc}, our approach performs substantially better than the standard FFNN approach in all cases. The differences between the two approaches in both closest-distance and zone-intrusion values are statistically significant upon the application of the t-test and z-test with a confidence level of 0.95.

\subsection{Analysis of the Local Controller: Evasive Maneuvers}

Our model makes instantaneous action decisions at the local-controller level to respond to the changes in the environment. For convenience, the model considers only the closest pedestrian, but the underlying neural network can handle multiple pedestrians. When it is inevitable to diverge from a straight path due to a possible conflict with a pedestrian, the local controller navigates the robot safely, following the norms taught by the expert demonstrations. To establish these claims, the performance of the trained local controller is assessed on a set of simulated tasks.

\begin{figure}[t]
  \begin{center}
    \begin{subfigure}{0.49\columnwidth}
        \includegraphics[width=\linewidth]{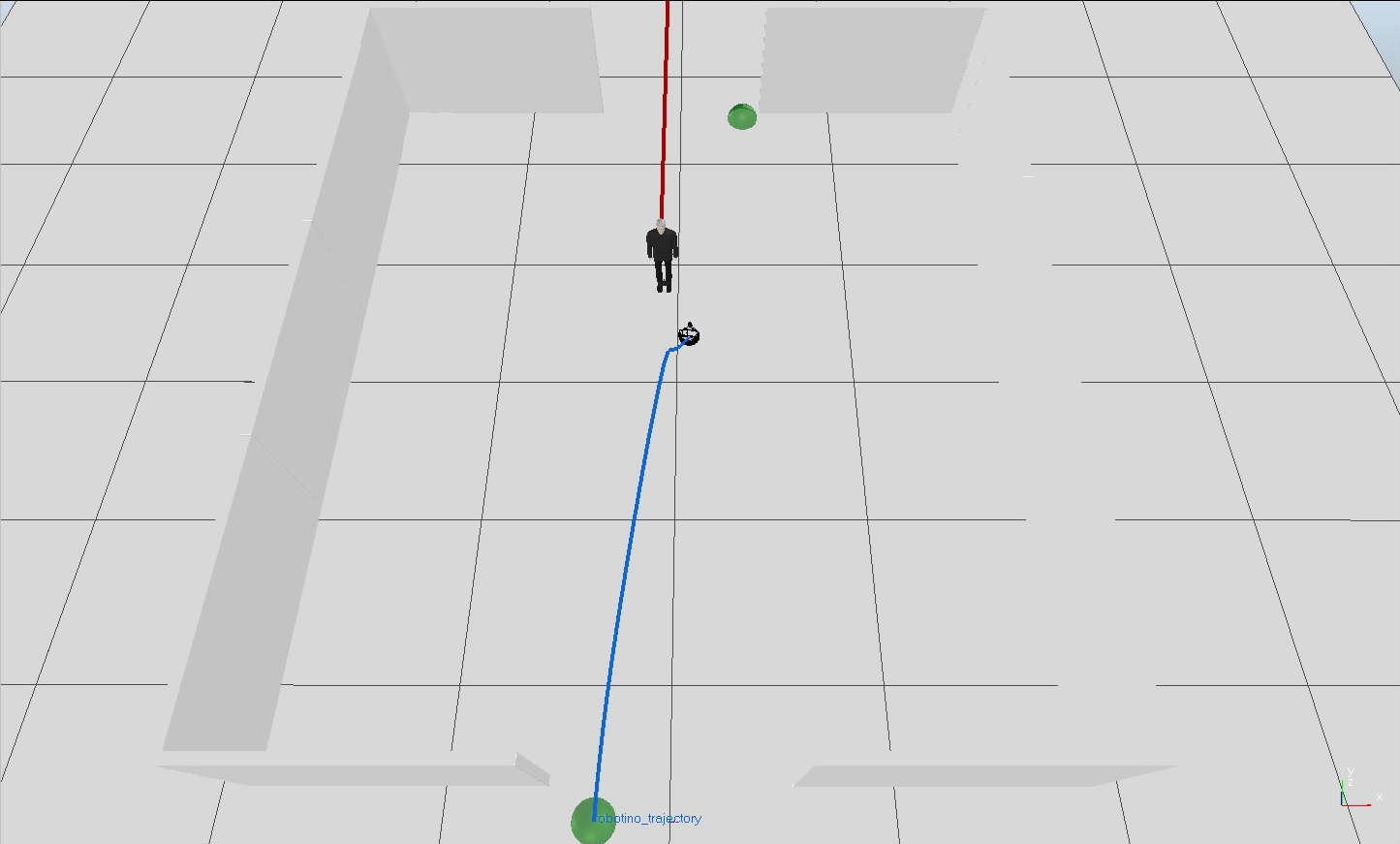}
    \end{subfigure}
    \hfill
    \begin{subfigure}{0.49\columnwidth}
        \includegraphics[width=\linewidth]{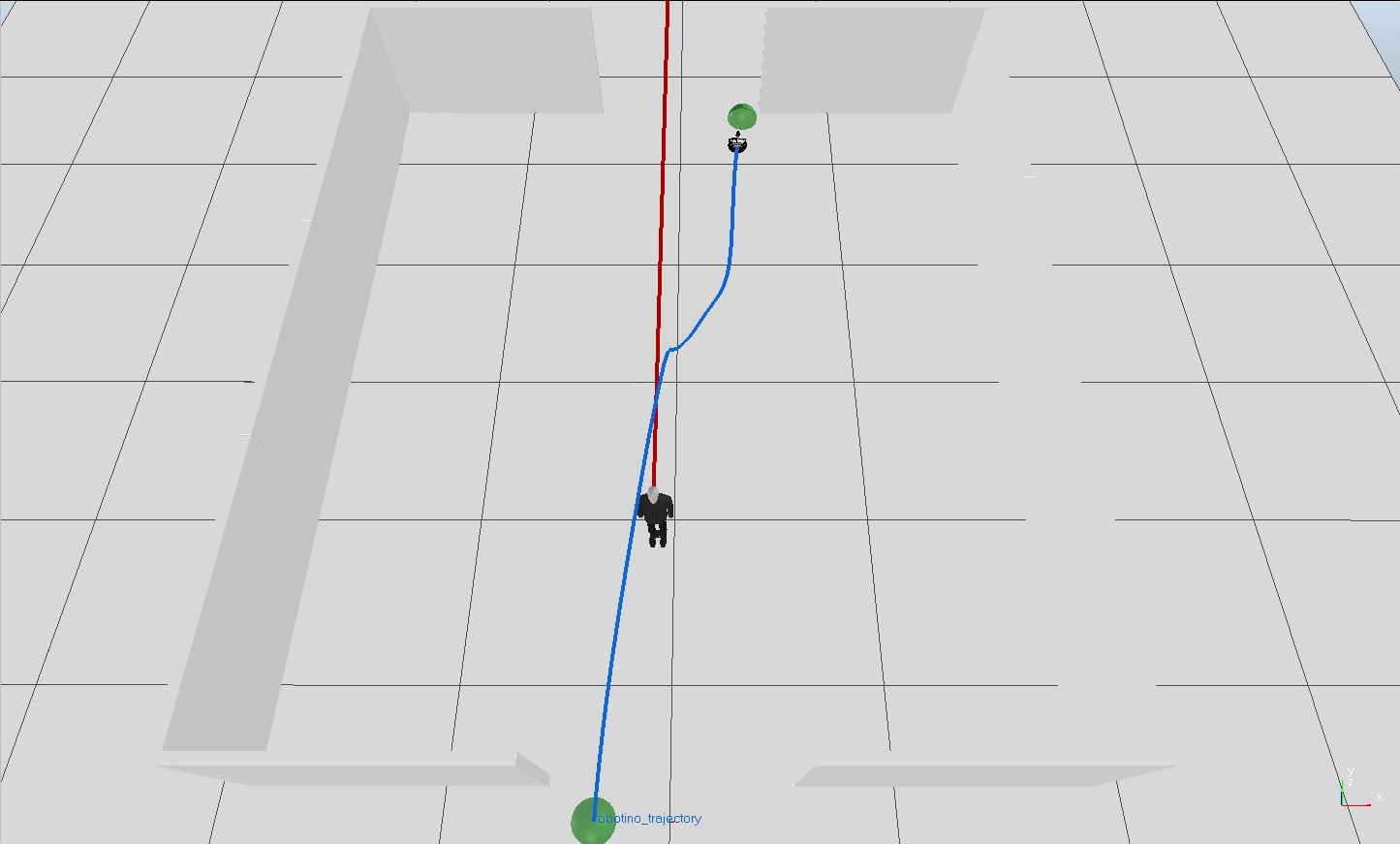}
    \end{subfigure}
    \caption{Snapshots from the scene where the pedestrian moves on a vertical trajectory shown in red. The robot, controlled by the Data-Driven Local Controller, aims to reach its target, shown at the top. When it encounters the pedestrian, it evades the pedestrian, reactively changing its trajectory, shown in blue.}
    \label{fig:exp_vert}
  \end{center}
\end{figure}

\begin{figure}[t]
  \begin{center}
    \begin{subfigure}{0.27\columnwidth}
        \includegraphics[width=\linewidth]{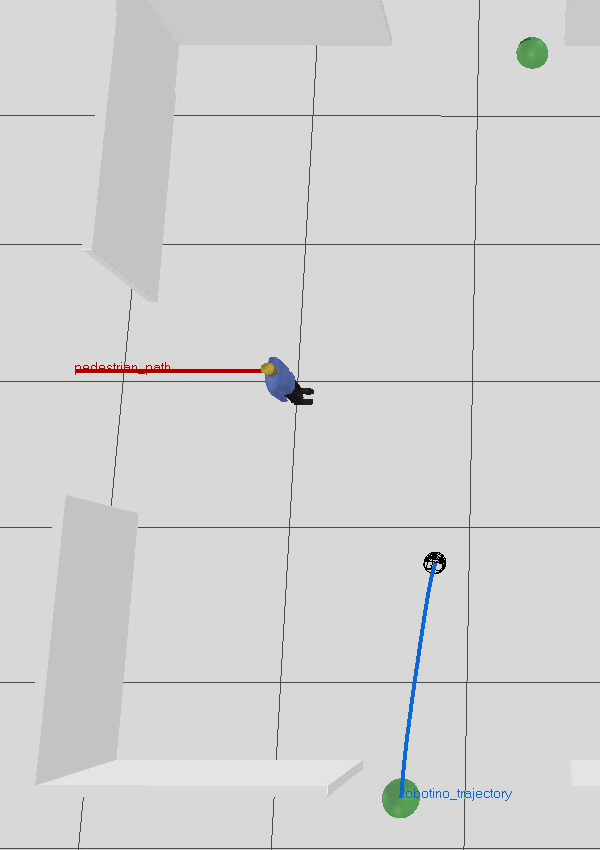}
        \caption{}
        \label{fig:exp_hora}
    \end{subfigure}
    \hfill
    \begin{subfigure}{0.27\columnwidth}
        \includegraphics[width=\linewidth]{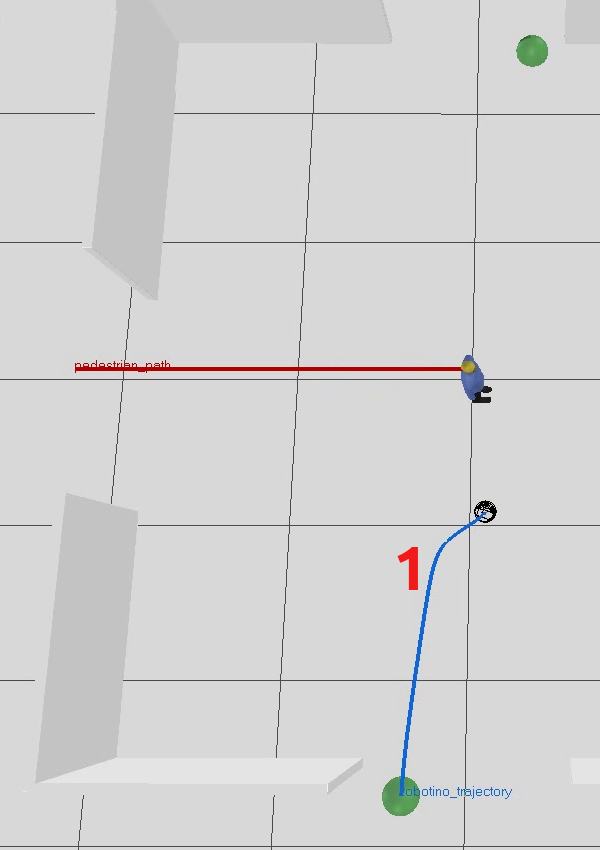}
        \caption{}
        \label{fig:exp_horb}
    \end{subfigure}
    \hfill
    \begin{subfigure}{0.4338\columnwidth}
        \includegraphics[width=\linewidth]{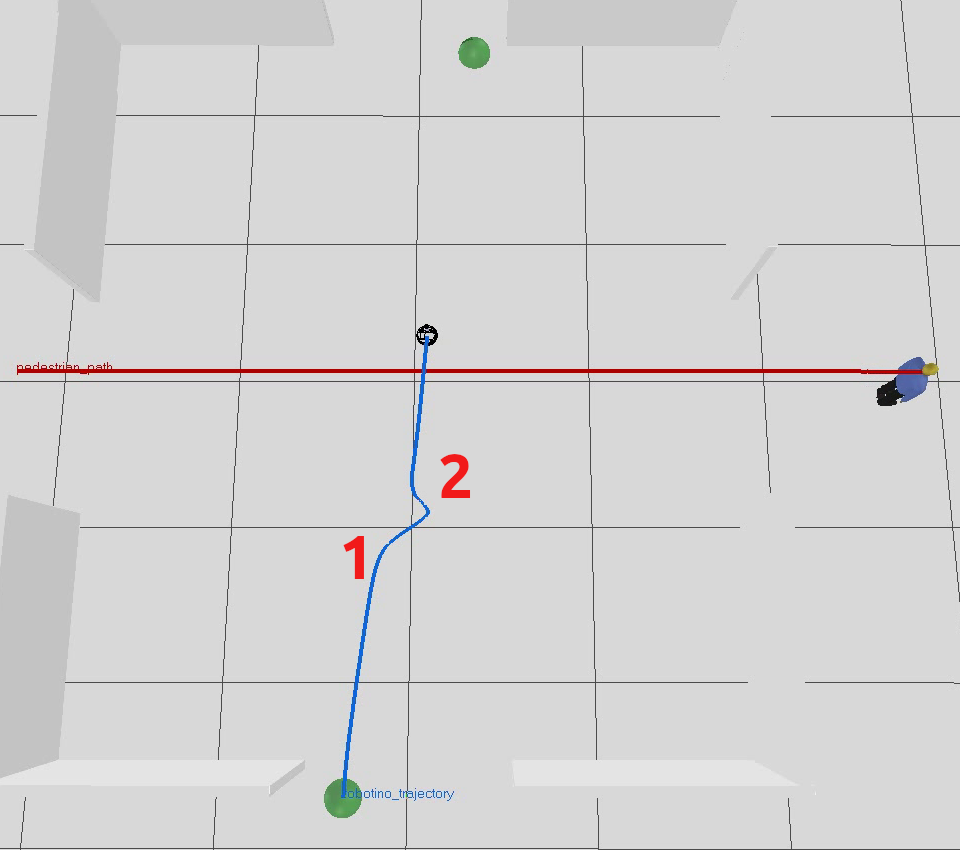}
        \caption{}
        \label{fig:exp_horc}
    \end{subfigure}
    \caption{Snapshots from the scene where the pedestrian is moving on a horizontal trajectory, shown in red whereas the trajectory of the robot is shown in blue. Figure \ref{fig:exp_hora} shows that the robot initially aims to reach its target, shown at the top with the green sphere. Figure \ref{fig:exp_horb} shows that it alters its trajectory first to evade the pedestrian on its left. As the pedestrian continues to move, Figure \ref{fig:exp_horc} shows that the robot again makes a maneuver to evade the pedestrian on its right.}
    \label{fig:exp_hor}
  \end{center}
\end{figure}

Firstly, the obstacle avoidance capability of the Data-Driven Local Controller was tested in configurations with a moving pedestrian. Although the training dataset does not contain any scenarios with moving pedestrians, our local controller could generalize to such cases. In Figure \ref{fig:exp_vert}, the robot alters its trajectory, shown in blue color, dynamically to steer away from the pedestrian, whose trajectory is shown in red color. The execution of this motion prior is more apparent in Figure \ref{fig:exp_hor}. In this case, the robot maneuvers multiple times to avoid any probable zone intrusions while moving toward its goal.

\begin{figure}
  \begin{center}
    \begin{subfigure}{0.49\columnwidth}
        \includegraphics[width=\linewidth]{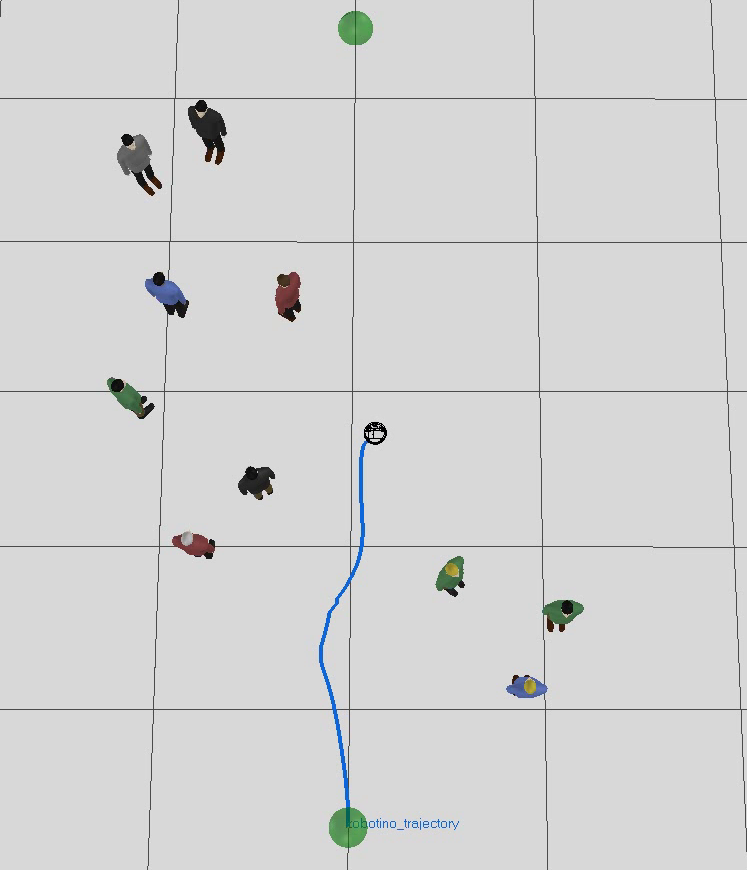}
    \end{subfigure}
    \hfill
    \begin{subfigure}{0.49\columnwidth}
        \includegraphics[width=\linewidth]{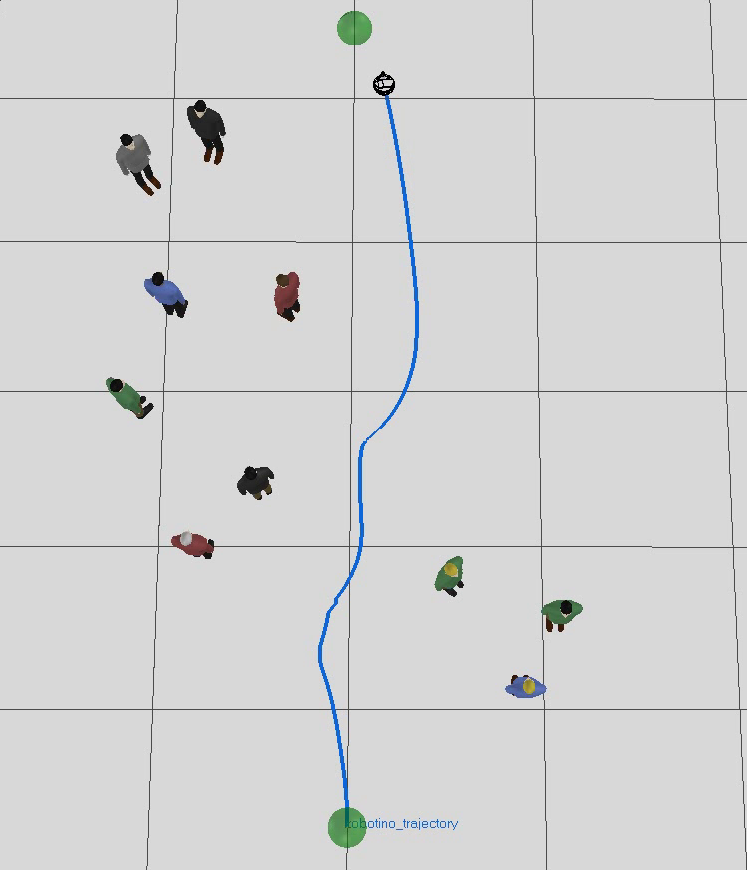}
    \end{subfigure}
    \caption{Snapshots from the scene where the robot is passing through several pedestrians. Instead of going directly to its target, the robot takes a longer trajectory (shown in blue), complying with the social norms present in the dataset. This behavior decreases proxemics-zone intrusions.}
    \label{fig:exp_mult}
  \end{center}
\end{figure}

\begin{figure}[h]
  \begin{center}
    \begin{subfigure}{0.75\linewidth}
        \includegraphics[width=0.75\linewidth]{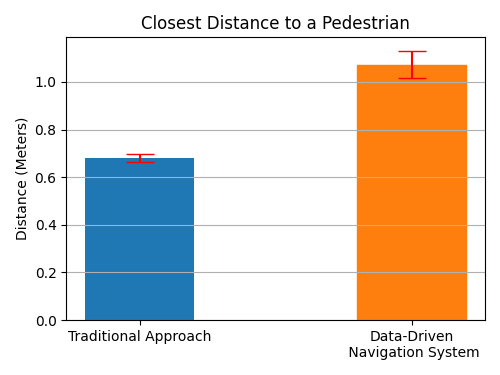}
        \caption{}
        \label{fig:cea}
    \end{subfigure}
    \vfill
    \begin{subfigure}{0.75\linewidth}
        \includegraphics[width=0.75\linewidth]{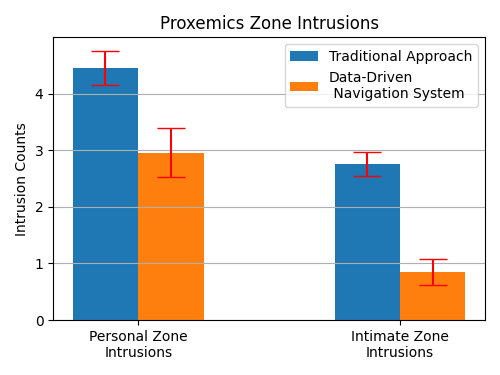}
        \caption{}
        \label{fig:ceb}
    \end{subfigure}
    \vfill
    \begin{subfigure}{0.75\linewidth}
        \includegraphics[width=0.75\linewidth]{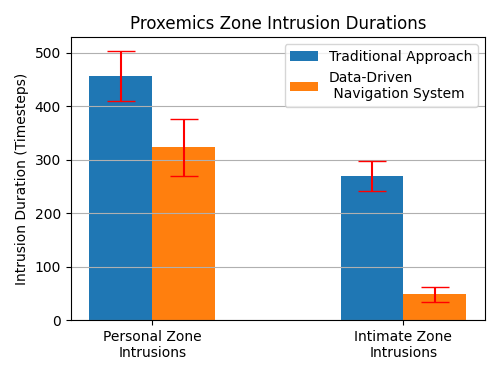}
        \caption{}
        \label{fig:cec}
    \end{subfigure}
    \caption{Comparison of approaches in terms of distance-based comfort metrics. In Figure \ref{fig:cea}, we show the closest distances to a pedestrian both approaches produce. Our pipeline minimizes the discomfort by staying distant from the pedestrians during navigation. This characteristic is also reflected in Figures \ref{fig:ceb} and \ref{fig:cec} where we compare two methods in terms of zone-intrusion metrics. The count and the duration of zone intrusions are significantly less for our system. Refer to the text for more details.}
    \label{fig:comp_entire}
  \end{center}
\end{figure}

Second, we tested the Data-Driven Local Controller in a scenario where multiple pedestrians were situated. Although the controller is trained in environments with single and stationary pedestrians, it can scale this behavior into more crowded environments by considering only the closest pedestrian at test time. In Figure \ref{fig:exp_mult}, the social aspect of the controller is rather apparent. Instead of going towards the goal directly, which is efficient and physically safe, the robot exhibits a social behavior of steering away from the pedestrians as its priority. Since such a norm was present in the expert behavior, the Data-Driven Local Controller captures this as a motion primitive.

\subsection{Comparison of Local Controllers}

To assess the success of our system, we devised a set of 25 simulated tests with three local controllers; Social Force Model, CNP trained on Simulation, and CNP trained on SCAND. SCAND  \citep{karnan2022socially} is a real-world dataset gathered on indoor and outdoor social environments with human demonstrators. More details on the dataset are given in Appendix \ref{app:dat}. The quantitative comparison is given in Table \ref{tab:comp3}.

\begin{table}[h]
\begin{tabular}{|c|c|c|c|}
\hline
              & \textbf{\begin{tabular}[c]{@{}c@{}}Social Force\\ Model\end{tabular}} & \textbf{\begin{tabular}[c]{@{}c@{}}CNP on \\ Simulation\end{tabular}} & \textbf{\begin{tabular}[c]{@{}c@{}}CNP on\\ SCAND\end{tabular}} \\ \hline
\textbf{ADE}  & 0.43 ± 0.01                                                           & 0.53 ± 0.01                                                           & 0.72 ± 0.02                                                     \\ \hline
\textbf{PL}   & 13.18 ± 0.01                                                          & 13.49 ± 0.03                                                          & 13.10 ± 0.01                                                    \\ \hline
\textbf{ATG}  & 22.83 ± 0.08                                                          & 34.83 ± 0.38                                                          & 69.86 ± 0.80                                                    \\ \hline
\textbf{AA}   & 5.24 ± 1.51                                                           & 6.51 ± 0.07                                                           & -0.06 ± 0.07                                                    \\ \hline
\textbf{PZIC} & 0                                                                     & 0                                                                     & 3                                                               \\ \hline
\end{tabular}
\caption[Comparison]{Three local controllers are compared in simulated tests on 25 environments on five dimensions. Average Displacement Error (ADE) shows how much the robot steers away from the straight line from the start position to the goal position. Path Length (PL) is used to measure the length of the generated trajectories whereas Average Time to Goal (ATG) shows how long it takes for the robot to get to the goal position. Average Acceleration (AA) is reported by summing up the changes in the acceleration throughout the navigation task. Proxemics Zone Intrusion Counts (PZIC) measure the number of personal zone intrusions. SI units are used in the comparison. Refer to the text for details.}
\label{tab:comp3}
\end{table}

Several quantitative metrics are used to compare the three controllers. Average Displacement Error (ADE) and Average Acceleration (AA) metrics are proposed in \cite{mavrogiannis2021core}; whereas Path Length (PL) and Average Time to Goal (ATG) in \cite{okal2016formalizing}. Proxemics Zone Intrusion Counts (PZIC) metric is proposed in \cite{vasquez2014inverse}. According to these tests, the most notable difference is the velocity each agent prefers in the same condition. Demonstrators seem to prefer lower speeds around people, whereas our simulated controller and the one trained on this data travel at higher speeds. Therefore, even though PL values are very close, the controller trained on real-world data scores substantially higher ATG values. This situation is also reflected in the AA dimension. The controller trained on real-world data tends to change its speed less frequently than the other two. On the other hand, this controller makes more personal zone intrusions, signaling that human demonstrators prefer slowly diverging from other pedestrians' paths in case of encounters.

\subsection{Performance of the Complete System}

Finally, we demonstrate the contribution of our pipeline in terms of social navigation with a quantitative comparison. In the same environment as shown in Figure \ref{fig:exp_mult}, we randomly distributed the pedestrians and compared the executed navigation trajectories of the proposed system and a traditional navigation system. The traditional planner implements the A* algorithm at the global level \citep{astar}, and the Elastic Bands approach at the local level \citep{quinlan1993elastic}. Figure \ref{fig:comp_entire} summarizes the numerical analysis. In Figure \ref{fig:cea}, we compare both methods in terms of the closest distances they approach a pedestrian. Closer distances negatively affect the comfort levels of pedestrians \citep{diego2011please}. The comparison shows that our approach leads to less discomfort by leaving more space with the surrounding people. In Figure \ref{fig:ceb}, we present proxemics-zone intrusion counts for both methods and in Figure \ref{fig:cec}, zone-intrusion durations are shown. These comparisons also show that our pipeline makes fewer and more brief intrusions of the proxemics zones of pedestrians. Lastly, in terms of distance-based metrics, these differences are statistically significant upon the application of the t-test with a confidence level of 0.95.

\begin{figure}[t]
  \begin{center}
    \begin{subfigure}{0.49\columnwidth}
        \includegraphics[width=\linewidth]{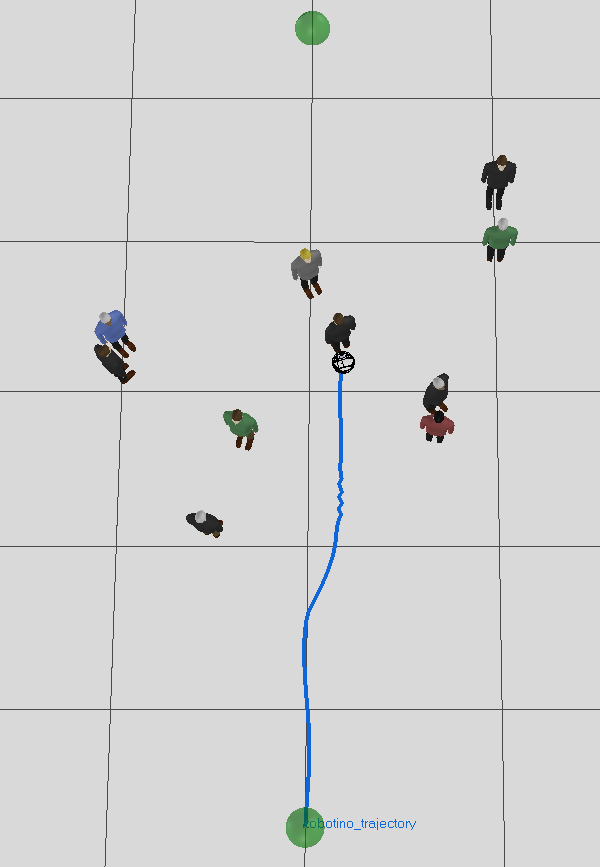}
        \caption{}
        \label{fig:wgana}
    \end{subfigure}
    \hfill
    \begin{subfigure}{0.49\columnwidth}
        \includegraphics[width=\linewidth]{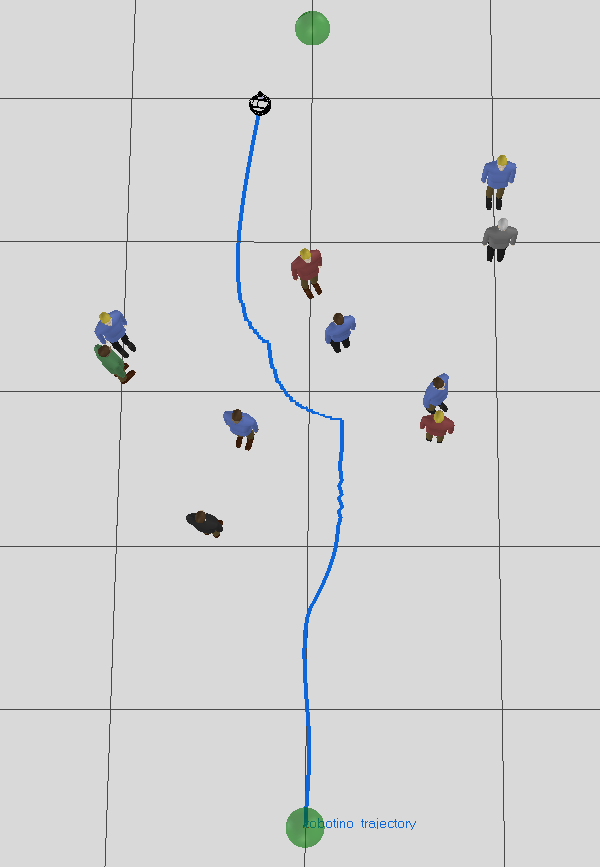}
        \caption{}
        \label{fig:wganb}
    \end{subfigure}
    \caption{Navigation system with and without the Failure Prediction Module in the same environment. Figure \ref{fig:wgana} shows when the robot relies only on the data-driven modules of the navigation system. This leads to an extrapolation error and eventually a collision. In Figure \ref{fig:wganb}, the Failure Prediction Module detects this situation and gives the control temporarily to the Hand-Crafted Reactive Controller, enabling the robot to move safely within the cluttered area.}
    \label{fig:wgan}
  \end{center}
\end{figure}

\subsection{Contribution of the Failure Prediction Module}

The Failure Prediction Module that we incorporated into our navigation system aims to recognize potential extrapolating cases before any collision occurs. In the following, we first illustrate the contribution of this module. Later, we present an ablation study where we measure the performance of the complete pipeline with and without the Failure Prediction Module.

Figure \ref{fig:wgana} shows a failed extrapolation of the data-driven modules. The Failure Prediction Module detects such extrapolation cases and enables the navigation system to take measures, as shown in Figure \ref{fig:wganb}.

\begin{figure}
  \begin{center}
    \begin{subfigure}{0.8\columnwidth}
        \includegraphics[width=\linewidth]{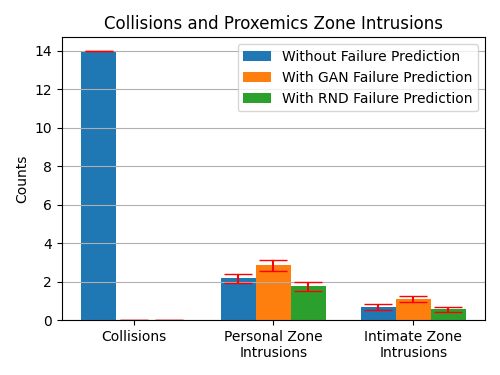}
    \end{subfigure}
    \caption{Performance of the Data-Driven Navigation System with and without the Failure Prediction Module in different environments with randomly placed pedestrians. Integration of the Failure Prediction Module successfully resolves collisions in each case. Refer to the text for more details.}
    \label{fig:cg_cwog}
  \end{center}
\end{figure}

In order to demonstrate the contribution of the Failure Prediction Module to the Data-Driven Navigation System, we compare the performance of the entire navigation system with and without this module. Also, we present a comparison between the two outlier detection approaches on our system, namely the GAN and the RND. We ran 50 tests with randomly positioned pedestrians in the environment, just like the one shown in Figure \ref{fig:wgan}. The numbers of collisions and proxemics-zone intrusions for three cases are shown in Figure \ref{fig:cg_cwog}.

Extrapolation errors occurred in 14 tests, and without the Failure Prediction Module such errors eventually led to collisions with pedestrians. On the other hand, the addition of the Failure Prediction Module enabled the successful prediction of all extrapolating cases with the exception on one case where RND notices the extrapolation too late for the robot to recover. Upon the prediction of extrapolation, the Hand-Crafted Reactive Controller Module temporarily took control of the mobile base, avoiding collisions altogether in all instances. However, results show that the numbers and durations of the proxemics-zone intrusions are higher for the system with the Failure Prediction Module. There are two reasons for this situation. First, tests are finished early when a collision occurs; the robot could simply not complete tests with collisions. Therefore, it was not possible to examine the performance in the rest of the task leading to a low number of intrusions. Second, although the system with the Failure Prediction Module successfully executed social navigation in most cluttered cases, some of these environments were just too cluttered. In these cases, the navigation pipeline eluded collisions by turning them into zone intrusions.

\subsection{Scalability of CNP}

To evaluate the scalability of our controllers, we train the Data-Driven Local Controller Module with datasets of different sizes and perform simulated tests with the learned controllers. The results are given in Table \ref{tab:cs}. According to these tests, when the dataset size is small, when the size is $D=50$ or $D=250$, we see that the controller struggles to avoid collisions with pedestrians. The controller starts to learn meaningful avoidance behaviors when $D \ge 500$. As \textit{D} increases, we see a decrease in the proxemics-zone intrusion counts. Average Displacement Error reaches meaningful levels only when $D > 1000$.

\begin{table}[h]
\begin{tabular}{|c|c|c|c|c|}
\hline
                  & \textbf{\begin{tabular}[c]{@{}c@{}}Collision\\ Count\end{tabular}} & \textbf{ADE} & \textbf{IZIC} & \textbf{PZIC} \\ \hline
\textbf{D = 50}   & 5                                                                  & 0.17 ± 0.02  & 5             & 5             \\  \hline
\textbf{D = 250}  & 4                                                                  & 0.3 ± 0.03   & 5             & 6             \\  \hline
\textbf{D = 500}  & 0                                                                  & 1.12 ± 0.02  & 5             & 10            \\   \hline
\textbf{D = 1000} & 0                                                                  & 0.25 ± 0.02  & 4             & 7             \\  \hline
\textbf{D = 1500} & 0                                                                  & 0.96 ± 0.02  & 0             & 0 \\ \hline

\end{tabular}
\caption[Comparison]{To evaluate how the size of the dataset affects the learning outcome of the Data-Driven Local Controller, we conduct a set of tests on simulation, with changing sizes (\textit{D}) of expert demonstrations.  Here, ADE refers to Average Displacement Error, while IZIC and PZIC refer to Intimate and Personal Zone Intrusion Counts. Please refer to the text for details.}
\label{tab:cs}
\end{table}

\section{Conclusion}

This study presents a novel navigation system that can learn local and global navigation directly from expert demonstrations. We used a state-of-the-art deep learning framework, namely Conditional Neural Processes (CNPs), that can learn multimodal distributions around complex trajectories from relatively smaller datasets compared to its alternatives. One common problem with data-driven systems is that they are prone to extrapolation errors. Therefore, using only the data-driven architectures would eventually lead our system into collisions. To minimize this possibility, we devise a layered architecture inspired by the subsumption architecture \citep{brooks1986robust}. We leverage state-of-the-art deep learning frameworks, namely Random Network Distillation (RND) and Generative Adversarial Networks (GANs), to work as arbitrators in this hierarchy. We have compared their performances and found that RND works better on average to detect out-of-distribution (OOD) samples enabling the system to infer the situations leading to extrapolation errors. When OOD states are encountered, the control of the robot is temporarily given to a manually encoded controller. We use an SFM-based controller for this purpose, but any socially-aware manually-encoded controller can be employed in the Hand-Crafted Reactive Controller.

In a simulated environment with a mobile robot, stationary and moving pedestrians, and given target points, we showed that our method could generate socially aware paths at the global level and successfully avoid pedestrians at the local level. This idea opposes the conventional approach of handling social navigation only in the local layer. An increase in the socialness of global trajectories improves the overall success of the social navigation frameworks. The results were verified in different simulated environments using metrics such as the average acceleration, path length, average displacement error, the closest distance to pedestrians and intrusion amount of their proxemics zones. These promising findings should be supported by real-world tests as reactions of the simulated people do not align perfectly with actual humans in natural environments. On the other hand, there are certain limitations of the current status of the system. An important limitation of this work is that the model does not consider the orientation and the velocity of the pedestrians. These factors would definitely alter the global and local plans of the robot. In future work, we plan to include these parameters in our models. Moreover, since we applied interpolation on the demonstration trajectories to get a continuous path, we lose the timing aspect of the trajectory in the Data Driven Global Controller Module. We are planning to solve this issue by adopting a different representation that enables us to learn global trajectories from the real-world data without interpolation so that we can produce more flexible global trajectories. There are recent benchmarking tools in the social robot navigation domain, such as \cite{biswas2022socnavbench, tsoi2020sean, holtz2021socialgym}. They provide an automated evaluation of the algorithms and comparison with several conventional approaches, such as SFM and RVO \citep{van2008reciprocal}. We plan to use these tools to automate and expedite benchmarking efforts.

We believe that the social aspects of our approach can further benefit from the integration trajectory forecasting mechanisms, such as \cite{gupta2018social}. In the future, we plan to use future predictions about pedestrian trajectories as input to the trajectory calculation of the robot to create less disturbing trajectories by increasing the preference for uncongested parts of the environment, avoiding probable future encounters. Our system uses CNPs in both global and local controller modules. On the other hand, these two modules are independent of each other. We believe we can leverage the capability of CNPs in both layers, but some different approaches, like LSTMs or transformers, can also be used here. We also aim to investigate such techniques in the global controller layer in the future. Moreover, we aim to deploy this system on real robots. Since robots in the real-world need to be controlled in real-time for successful navigation, the challenge in applying our system is obtaining high-level parameters that we feed to CNPs in real-time. We believe that this can be addressed by leveraging the flexibility of CNPs to learn representations from any low-level and high-dimensional input, as shown in \cite{gordon2019convolutional}.

\section{Acknowledgements}
This work was supported by the BAGEP Award of the Science Academy. Thanks to Alper Ahmetoglu for helping with the implementation.

\bibliography{ref}

\appendix

\include{appendix}

\end{document}

%% file: appendix.tex
\section{Datasets}
\label{app:dat}

\subsection{Real-World Data}
We trained the model on a real-world dataset, namely SCAND \citep{karnan2022socially}. This dataset contains more than 8 hours of navigation trajectories in social environments, recorded using a teleoperated robot. Along these trajectories, researchers also present the raw data from several sensors, including cameras, lidars, GPS, etc.

\begin{figure}[h]
  \begin{center}
    \begin{subfigure}{0.49\columnwidth}
        \includegraphics[width=\columnwidth]{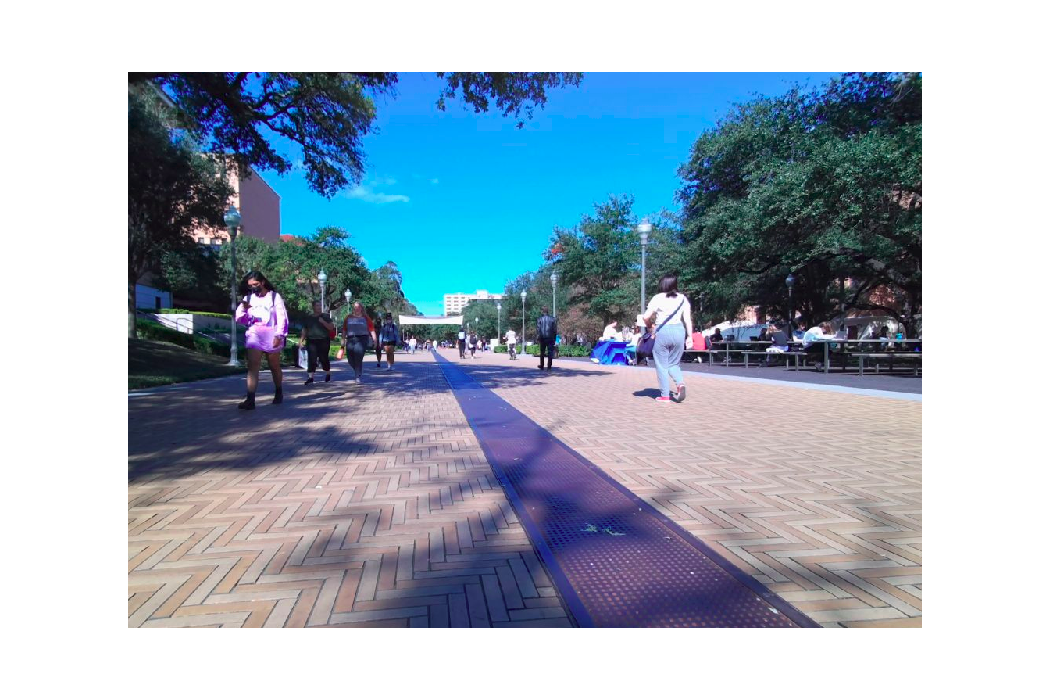}
    \end{subfigure}
    \hfill
    \begin{subfigure}{0.49\columnwidth}
        \includegraphics[width=\columnwidth]{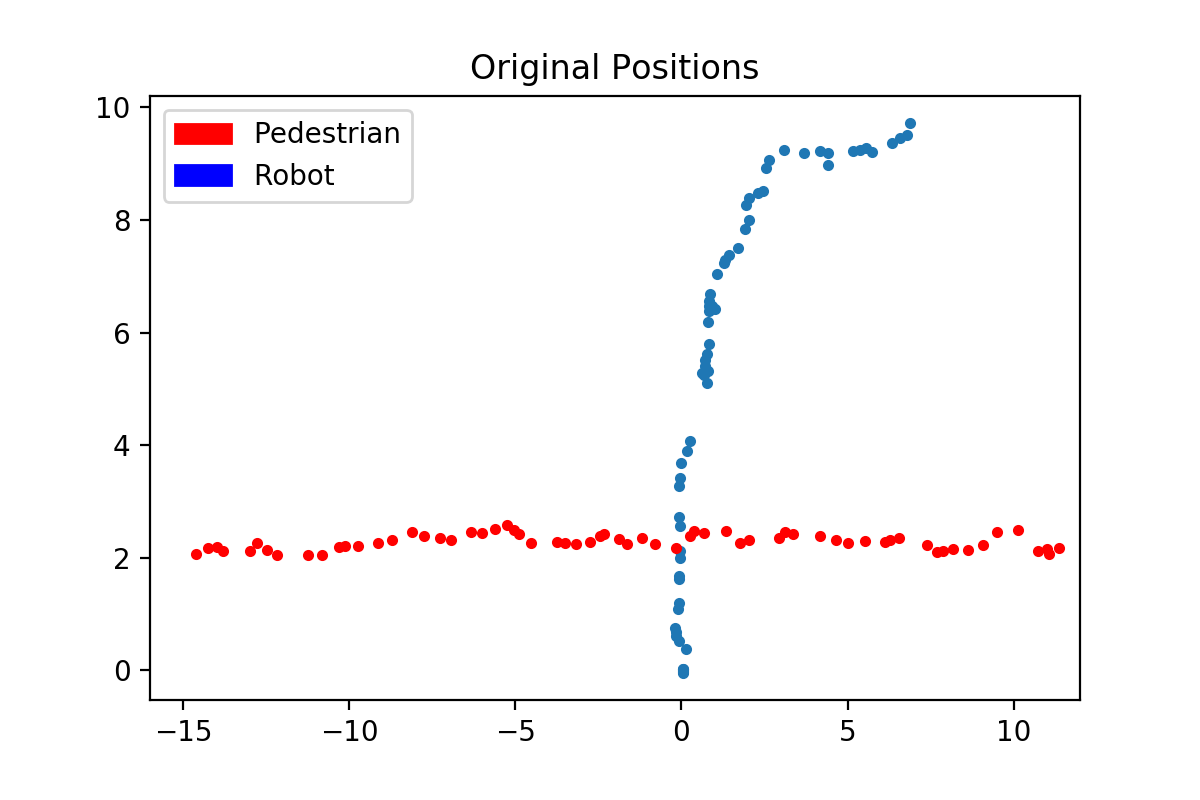}
    \end{subfigure}
    \vfill
    \begin{subfigure}{0.49\columnwidth}
        \includegraphics[width=\columnwidth]{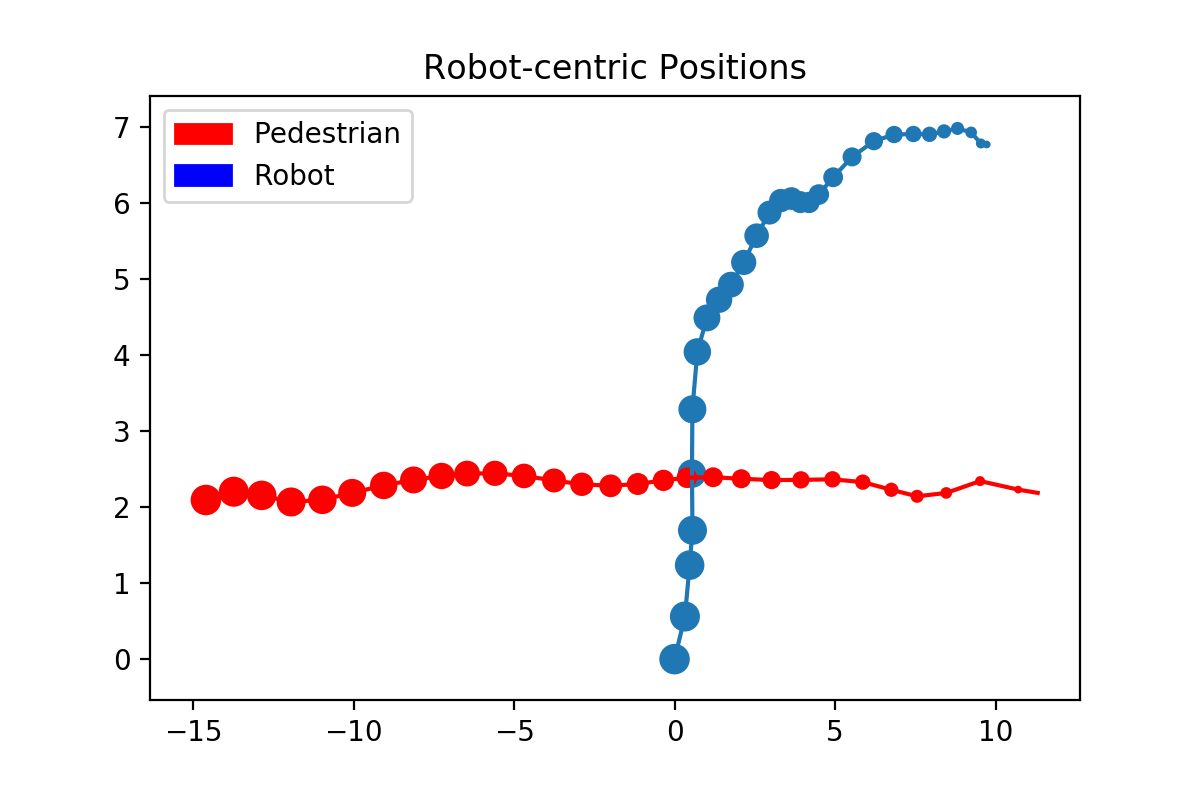}
    \end{subfigure}
    \hfill
    \begin{subfigure}{0.49\columnwidth}
        \includegraphics[width=\columnwidth]{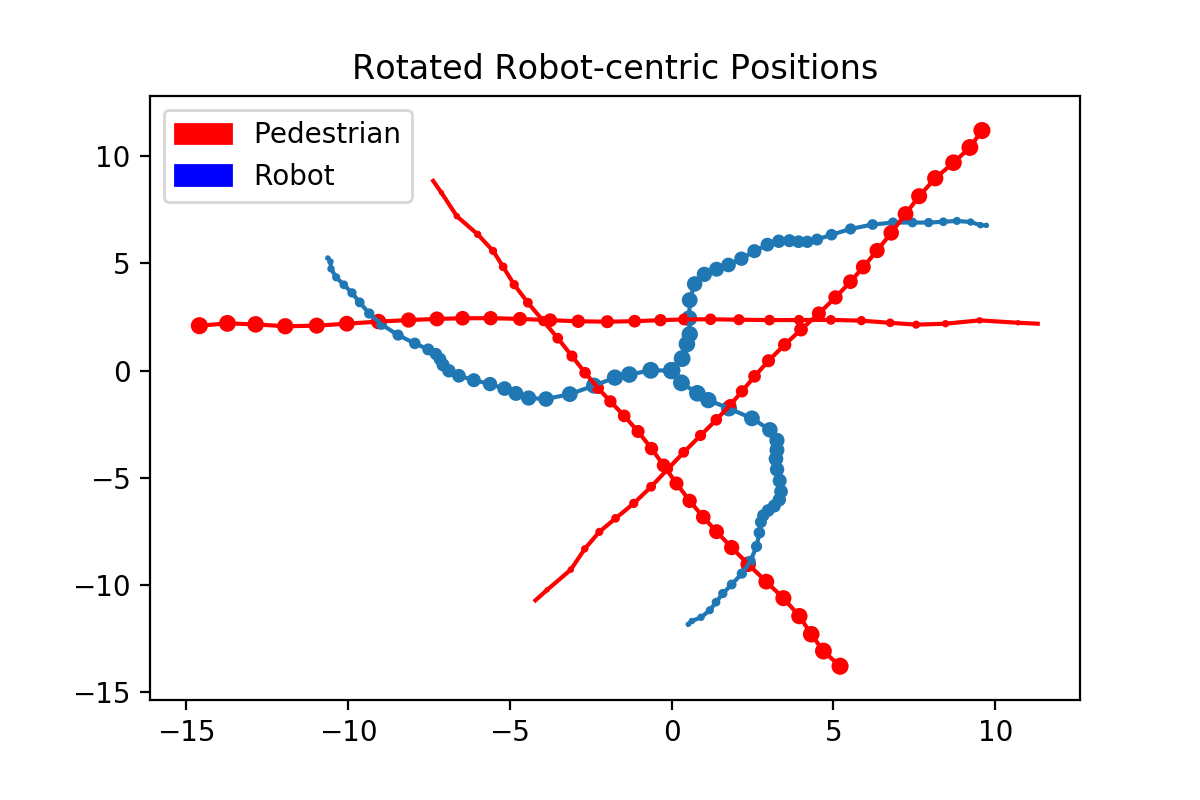}
    \end{subfigure}
    \caption{Processing the data from SCAND \citep{karnan2022socially}. First, raw sensor data of the selected trajectory clips with evasive maneuvers are gathered. Robot and pedestrian positions are annotated on the data and transformed into a robot-centric coordinate frame. Then, interpolation is applied to obtain entire trajectories. Finally, rotation is applied to increase the size of the dataset.}
    \label{fig:scand}
  \end{center}
\end{figure}

In this dataset, first, we have manually extracted the navigation data of 30 social scenarios where the robot encounters a pedestrian and avoids collision by maneuvering. In each encounter, we manually annotated the positions of the pedestrians at multiple time steps. At the same time, we obtained the robot's positions using the GPS sensor and recorded this information with corresponding time stamps. Later, we applied interpolation to position data to get the robot's and pedestrians' continuous trajectories. As in \cite{perez2018learning}, where researchers use robot-centric data to learn social navigation, we processed the position data of timed trajectories to convert it to egocentric. Then, we applied rotational transformations to egocentric trajectories for data augmentation purposes. Data augmentation is the process of increasing the size of the dataset artificially. This technique has been used in many deep learning applications widely \citep{shorten2019survey}. The entire data-processing procedure is depicted in Figure \ref{fig:scand}.

\subsection{Simulated Data}
\begin{figure}[h]
  \begin{center}
    \begin{subfigure}{0.49\columnwidth}
        \includegraphics[width=\linewidth]{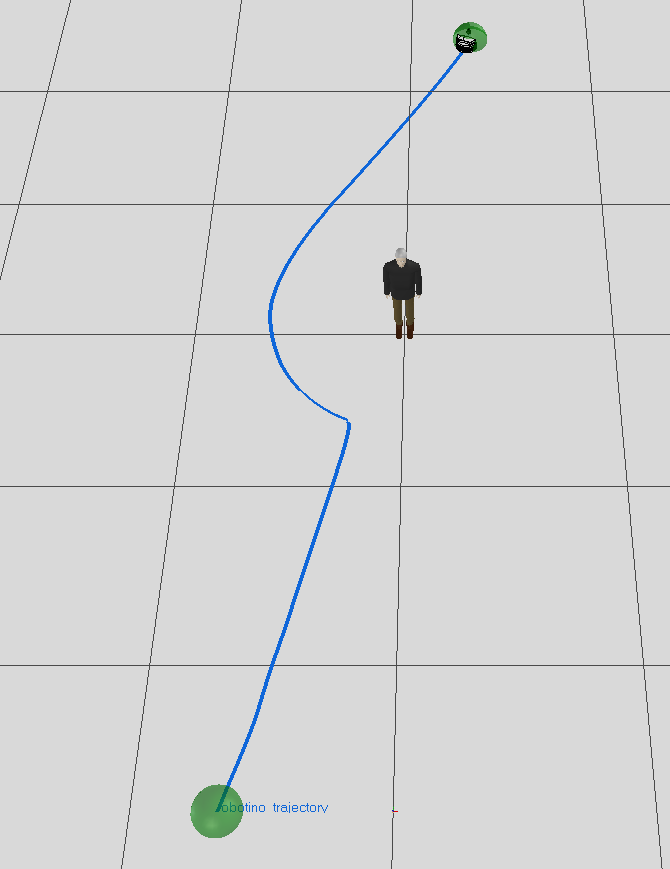}
    \end{subfigure}
    \hfill
    \begin{subfigure}{0.49\columnwidth}
        \includegraphics[width=\linewidth]{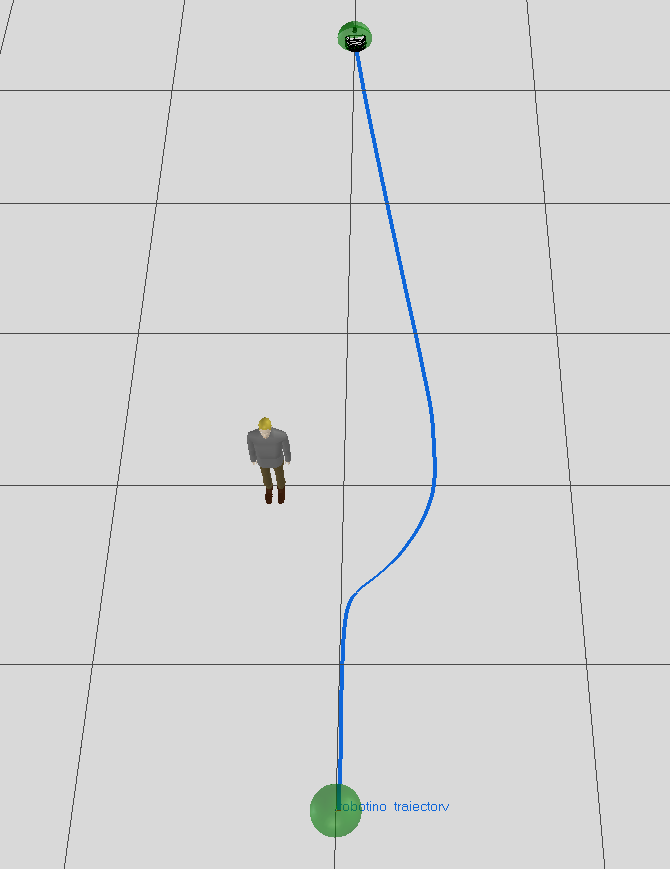}
    \end{subfigure}
    \caption{Data collection on the simulation. The robot implements Social Force Model to avoid randomly placed pedestrians. Green spheres show the navigation task's start and goal positions, and the blue line shows the motion trajectory.}
    \label{fig:dc} 
  \end{center}
\end{figure}

To gather a set of socially-compliant demonstration trajectories for learning, we implemented the Social Force Model, described in \cite{helbing1995social} to control the robot. Assuming that it generates socially plausible trajectories, we recorded 1500 trajectories with random start, goal, and pedestrian positions. In each trial, single and multiple stationary and dynamic pedestrians are placed at random positions. Figure \ref{fig:dc} shows snapshots of this procedure from the simulator.

%%%%%%%%%%%%%%%%%%%%%%%%%%%%%%%%%%%%%%%%%%%%%%%%%%%%%%%%%%%%%%%%%%%%%%%%%

\section{Models}
\label{app:mod}

The implementation of the model and the training data can be found at \url{https://github.com/yildirimyigit/cnmp}. In addition, data-processing scripts, ROS nodes, and environments used in the simulator can be found at \url{https://github.com/yildirimyigit/irl_sfm}.

In the following, we present the control parameters of all models used throughout the study.

\newpage

\subsection{Social Force Model}

\begin{table}[h]
\begin{tabular}{|l|l|}
\hline
\textbf{Hyperparameter} & \textbf{Value} \\ \hline
Relaxation Time         & 2.3            \\ 
Force Strength          & 6.40           \\ 
Force Range             & 0.25           \\ 
Maximum Velocity        & 2.5            \\ \hline
\end{tabular}
\vskip\baselineskip 
\caption[Parameters]{Model parameters of SFM}
\end{table}

\subsection{Data-Driven Global Controller (Conditional Neural Process)}

\begin{table}[h]
\begin{tabular}{|l|l|}
\hline
\textbf{Hyperparameter}                   & \textbf{Value} \\ \hline
Maximum Number of Observations            & 10             \\
Number of Hidden Layers                   & 3              \\
Width of Layers of Encoder Network        & 256, 256, 256  \\
Width of Layers of Query Network          & 256, 256, 256  \\
Optimizer                                 & Adam           \\
Activation Function                       & ReLU           \\
Learning Rate                             & 1e-4           \\ \hline
\end{tabular}
\vskip\baselineskip 
\caption[Hyperparameters]{Hyperparameters of the CNP of the Data-Driven Global Controller}
\end{table}

\subsection{Data-Driven Local Controller  (Conditional Neural Process)}
\begin{table}[h]
\begin{tabular}{|l|l|}
\hline
\textbf{Hyperparameter}                   & \textbf{Value} \\ \hline
Maximum Number of Observations            & 20             \\
Number of Hidden Layers                   & 3              \\
Width of Layers of Encoder Network        & 256, 384, 512  \\
Width of Layers of Query Network          & 512, 384, 256  \\
Optimizer                                 & Adam           \\
Activation Function                       & ReLU           \\
Learning Rate                             & 1e-4           \\ \hline
\end{tabular}
\vskip\baselineskip 
\caption[Hyperparameters]{Hyperparameters of the CNP of the Data-Driven Local Controller}
\end{table}

\vfill\eject

\subsection{Feed-Forward Neural Network}
Feed-Forward Neural Network is used for performance comparison with CNP.

\begin{table}[h]
\begin{tabular}{|l|l|}
\hline
\textbf{Hyperparameter} & \textbf{Value} \\ \hline
Number of Hidden Layers & 3              \\
Width of Hidden Layers  & 256, 512, 1024 \\
Optimizer               & SGD            \\
Activation Function     & ReLU           \\
Learning Rate           & 1e-3           \\ \hline
\end{tabular}
\vskip\baselineskip 
\caption[Hyperparameters]{Hyperparameters of the Feed-Forward Neural Network as the global controller}
\end{table}

\subsection{Failure Prediction Module - GAN}
\begin{table}[h]
\begin{tabular}{|l|l|}
\hline
\textbf{Hyperparameter} & \textbf{Value}                                                              \\ \hline
Number of Hidden Layers & 2                                                                           \\
Width of Hidden Layers  & 128, 128                                                                    \\
Generator Noise         & \begin{tabular}[c]{@{}l@{}}64 Dimensional,\\ Diagonal Gaussian\end{tabular} \\
Optimizer               & Adam                                                                        \\
Activation Function     & ReLU                                                                        \\
Learning Rate           & 1e-4                                                                        \\ \hline
\end{tabular}
\vskip\baselineskip 
\caption[Hyperparameters]{Hyperparameters used in the GAN of Failure Prediction Module}
\end{table}

\subsection{Failure Prediction Module - RND}
\begin{table}[h]
\begin{tabular}{|l|l|l|}
\hline
\textbf{Hyperparameter} & \textbf{\begin{tabular}[c]{@{}l@{}}Value in\\ Target\end{tabular}} & \textbf{\begin{tabular}[c]{@{}l@{}}Value in\\ Predictor\end{tabular}} \\ \hline
Number of Hidden Layers & 3                                                                  & 3                                                                     \\
Width of Hidden Layers  & 512,512,512                                                        & 512,512,512                                                           \\
Optimizer               & N/A                                                                & Adam                                                                  \\
Activation Function     & ELU                                                                & ELU                                                                   \\
Learning Rate           & N/A                                                                & 1e-4                                                                  \\ \hline
\end{tabular}
\vskip\baselineskip 
\caption[Hyperparameters]{Hyperparameters used in the RND of Failure Prediction Module}
\end{table}